\documentclass[12pt]{article}
\usepackage{amsmath,color,amsthm,amssymb,pdfpages,bbm}
\usepackage{graphicx}
\usepackage{algorithm,algorithmic}
\usepackage{caption}
\usepackage{subcaption}

\numberwithin{equation}{section}

\newcommand{\gB}{D}
\newtheorem{theorem}{Theorem}[section]
\newtheorem{corollary}[theorem]{Corollary}
\newtheorem{lemma}[theorem]{Lemma}
\newtheorem{proposition}[theorem]{Proposition}
\newtheorem{claim}[theorem]{Claim}
\newtheorem{example}[theorem]{\sl Example}
\newtheorem{definition}[theorem]{Definition}

\theoremstyle{definition}
\newtheorem{remark}{Remark}
\numberwithin{remark}{subsection}

\newcommand{\argmin}{\operatornamewithlimits{argmin}}

\newcommand{\hX}{\widehat{X}}
\newcommand{\hY}{\widehat{Y}}

\newcommand{\begp}{\begin{proposition}}
\newcommand{\enp}{\end{proposition}}
\newcommand{\begt}{\begin{theorem}}
\newcommand{\ent}{\end{theorem}}
\newcommand{\begl}{\begin{lemma}}
\newcommand{\enl}{\end{lemma}}
\newcommand{\begc}{\begin{corollary}}
\newcommand{\enc}{\end{corollary}}
\newcommand{\begcl}{\begin{claim}}
\newcommand{\encl}{\end{claim}}
\newcommand{\begr}{\begin{remark}}
\newcommand{\enr}{\end{remark}}
\newcommand{\begal}{\begin{algorithm}}
\newcommand{\enal}{\end{algorithm}}
\newcommand{\begd}{\begin{definition}}
\newcommand{\enf}{\end{definition}}
\newcommand{\begx}{\begin{example}}
\newcommand{\enx}{\end{example}}
\newcommand{\bega}{\begin{array}}
\newcommand{\ena}{\end{array}}

\newcommand{\ignore}[1]{}

\def\rompar(#1){\textup(#1\textup)}    

\begin{document}
\title{Spectral Clustering for Divide-and-Conquer Graph Matching}
\author{Vince Lyzinski$^{1}$, Daniel L. Sussman$^{2}$, Donniell E. Fishkind$^{3}$,\\ Henry Pao$^3$, Li Chen$^3$, Joshua T. Vogelstein$^{4}$,\\ Youngser Park$^{3}$, Carey E. Priebe$^{3}$\\
\small{$^1$ Human Language Technology Center of Excellence, Johns Hopkins University}\\
\small{$^2$ Department of Statistics, Harvard University}\\
\small{$^3$ Department of Applied Mathematics and Statistics, Johns Hopkins University}\\
\small{$^4$ Department of Biomedical Engineering, Johns Hopkins University}}
\date{}
\maketitle


\begin{abstract}
We present a parallelized bijective graph matching algorithm that leverages seeds and is designed to match very large graphs.  Our algorithm combines spectral graph embedding with existing state-of-the-art seeded graph matching procedures.  We justify our approach by proving that modestly correlated, large stochastic block model random graphs are correctly matched utilizing very few seeds through our divide-and-conquer procedure.
We also demonstrate the effectiveness of our approach in matching very large graphs in simulated and real data examples, showing up to a factor of 8 improvement in runtime  with minimal sacrifice in accuracy. 
\end{abstract}

\section{Introduction}
Graph matching is an increasingly important problem in inferential graph statistics, with applications across a broad spectrum of fields including computer vision (\cite{zhou}, \cite{cho2}), shape matching and object recognition (\cite {berg}, \cite{caelli}), and biology and neuroscience (\cite{haris}, \cite{FAQ}, \cite{zas}), to name a few.  The \emph{graph matching problem}  (GMP) seeks to find an alignment between the vertex sets of two graphs that best preserves common structure across graphs. 
Unfortunately, the GMP is inherently combinatorial, and
no efficient exact graph matching algorithms are known. Indeed, even the simpler problem of determining if two graphs are isomorphic is famously of unknown complexity (\cite{gj}, \cite{ssgt}), and if the graphs are allowed to be loopy, weighted and directed, then the simplest version of GMP is equivalent to the NP-hard quadratic assignment problem.  
Due to its wide applicability, there exist a vast number of approximating algorithms for GMP; see the paper ``30 Years of Graph Matching in Pattern Recognition" (\cite{30ygm}) for an excellent survey of the existing literature.   

When matching across graphs, often we have access to a partial matching of the vertices in the form of a {\it seeding}.  In practice, the assumption of seeds is quite natural in  many applications.  For example, in aligning social networks actors' user names may often allow for a partial alignment to be known a priori.  When matching across brain graphs (connectomes), we have geometric information provided by the brain atlas which provides a soft seeding fo the vertices.  In many time series graphs, it is common to have a group of invariant vertices across time which act as seeds.

In the {\it Seeded Graph Matching Problem} (SGMP), we leverage the information contained in an available partial matching to match the remaining vertices across graphs.  
Though the literature on seeded graph matching is comparatively small, recent results point to significant performance improvements in GM algorithms by incorporating even a modest number of seeds (\cite{FAP}, \cite{sgm2}).

Though a myriad of approximate graph matching algorithms exist, the very large graphs arising in the burgeoning realm of ``big data" demand highly scalable algorithms.  Roughly speaking, existing state of the art algorithms for approximate graph matching can be divided into two classes:  those that seek to bijectively match vertices of graphs of the same order, and those that seek matchings between the vertex sets that are allowed to be many--to--many and many--to--one.
The current cutting-edge bijective graph matching algorithms achieve excellent performance in approximately matching graphs with thousands of vertices and with computational complexity $O(n^3)$---$n$ the number of vertices being matched; 
see for example \cite{FAQ}, \cite{path} and \cite{jovo}.  These algorithms often operate directly on the adjacency matrices of the graphs to be matched, utilizing the tools of nonlinear optimization to approximtely solve GMP directly.  
However, owing to their $O(n^3)$ complexity, these algorithms are practically unusable, without significant computation resources, for matching very large graphs ($n\approx 10^5)$.  

Scalability is often achieved via relaxing the bijection requirement and allowing many--to--many and many--to--one matchings.  
These graph matching procedures can efficiently match very large graphs, often with $n$ in the tens of thousands; see for example \cite{knoss}, \cite{armiti}.  
A common approach to these scalable inexact algorithms is that they first match smaller, lower dimensional representative objects (prototype graphs in \cite{armiti}, eigenvectors in \cite{knoss}) and use these to build the overall matching.

Herein, we propose a new divide-and-conquer approach to {\it scalable bijective} seeded graph matching.  Our algorithm, the Large Seeded Graph Matching algorithm (LSGM, see Algorithm \ref{alg:dandc}), merges the approaches of bijective and non-bijective graph matching and leverages the information in seeded vertices in order to match very large graphs.
The algorithm proceeds in two steps: 
We first spectrally embed the graphs---yielding a low dimensional Euclidean representation of our graph---and then use the information provided by seeded vertices to jointly cluster the vertices of the two embedded graphs.   This embedding procedure allows us to employ the powerful theory of adjacency spectral embedding (see for example \cite{stfp} and \cite{fstp}) to prove asymptotically perfect performance in {\it jointly} clustering stochastic block model random graphs, see Theorem \ref{T} for detail. 

Once the vertices are jointly clustered, we then match the graphs within the clusters.   This matching step is fully parallelizable and flexible in that we can employ any one of a number of matching procedures depending on the properties of the resulting clusters.  
The flexibility afforded by our procedure in the clustering and matching subroutines can have a dramatic impact on algorithmic scalability.  For example, on a 1600 vertex simulated graph our parallelization procedure was able to achieve an  factor of 8 improvement in speed at minimal accuracy degradation by increasing the number of clusters and hence the number of cores that were used; see section \ref{sec:k}.

 Though we are not the first to employ a divide-and-conquer approach to graph matching (see for example \cite{cho}, \cite{zhou}, \cite{armiti}), our focus on the efficient utilization of apriori observed seeded vertices and the theoretical framework for our approach provided by Theorem \ref{T} set our algorithm apart from the existing literature. 


\noindent {\bf Note:} All graphs considered herein will be simple; in particular there are no multiple edges between two vertices nor are there edges with a single vertex as both endpoints.  
Modifications for the directed case are quite simple \cite{stfp,fstp} but we do not consider them in this manuscript.
All vectors considered will be column vectors, and $\vec{1}_m$ is the length-$m$ vector of all $1's$.  
When appropriate we drop the subscript and just write $\vec{1}$.
Throughout the paper we employ the standard notation $[n]:=\{1,2,\ldots,n\}$ for any $n\in\mathbb{N}$, and to simplify future notation, if $A\in\mathbb{R}^{n\times n}$ and $\tau, \sigma\subset [n]$, then $A(\tau,\sigma)$ will denote the submatrix of $A$ with row indices $\tau$ and column indices $\sigma$.  
For a matrix $X$, $X(:,i)$ will denote the $i$th column of $X$ and $X(i,:)$ the $i$th row of $X$.  Also for two matrices $X$ and $Y$, $[X|Y]$ will denote the column concatenation of $X$ and $Y$.

 \begin{algorithm}[t!]
\caption{Divide-and-conquer seeded graph matching; the LSGM algorithm}          
\label{alg:dandc}                           
\begin{algorithmic}
    \STATE{\bf INPUT:} Symmetric, hollow $A, B \in\{0,1\}^{n\times n}$, $s \in [n]$, seeding $\phi:[s]\rightarrow[s]$
    \STATE{\bf OUTPUT:} A matching of $G_1$ and $G_2$ given by $\psi$; 
\STATE{\bf Step 1:} Embed and jointly cluster the graphs according to Algorithm \ref{alg:emcl}  
\STATE {\bf Step 2:}  In parallel
\FOR{$i=1$ to $k$}  
\STATE Match cluster $i$ across the graphs using, yielding matching $\psi^{(i)}$;
\ENDFOR
\STATE {\bf OUTPUT:} $\psi= \oplus_{i=1}^k \psi^{(i)}$.
\end{algorithmic}
\end{algorithm}

\section{Background}
\label{S:B}
There are numerous formulations of the graph matching problem, though they all share the same objective heuristic:  given two graphs, $G_1=(V_1,E_1)$ and $G_2=(V_2,E_2)$, GMP seeks an alignment between the vertex sets $V_1$ and $V_2$ that best preserves structure across the graphs.  
In {\it bijective} graph matching, we further assume $|V_1|=|V_2|=n,$ and the alignment sought by GMP is a bijection between $V_1$ and $V_2$.  In {\it non-bijective} graph matching, we allow for $|V_1|\neq|V_2|$ and for alignments that are not one--to--one. 

In the bijective matching setting, GMP is commonly formulated as follows: find a bijection $\psi:V_1\rightarrow V_2$ minimizing the quantity
\begin{align} 
\label{eq:min}
\hspace{-4mm}\big|\big\{\,(i,j)\in V_1\times V_1\text{ s.t. }[i\sim_{G_1}\!j\text{, }\psi(i)\nsim_{G_2}\psi(j)]
\text{ or }
[i\nsim_{G_1}\!j\text{, }\psi(i)\sim_{G_2}\psi(j)]\big\}\big|,
\end{align}
i.e. the problem seeks to minimize the number of edge disagreements between $G_2$ and $``\psi(G_1)"$ (see \cite{FAQ}, \cite{path}, \cite{jovo}).  Equivalently stated, if $A$ and $B$ are the respective adjacency matrices of $G_1$ and $G_2$,
then this problem seeks to minimize 
$\|A-PBP^T\|_F^2,$ over all permutation matrices $P\!\in\!\Pi(n):=\{n\times n$ permutation matrices$\}$, with $\|\cdot\|_F$ the matrix Frobenius norm.  
In the non-bijective matching setting, $V_1$ and $V_2$ need not have equal cardinality.  This requires an alternative formulation of GMP, as (\ref{eq:min}) is no longer necessarily well-defined.  See \cite{caelli}, \cite{davies}, \cite{knoss}, \cite{zhou} for a variety of generalizations of (\ref{eq:min}). 


In the seeded graph matching problem (SGMP), we further assume the presence of a latent alignment $\phi$ between the vertex sets of $G_1$ and $G_2$.  Our task is to then efficiently leverage the information in a partial observation of the latent alignment, i.e. a \emph{seeding}, to estimate the remaining latent alignment.   In bijective SGMP, we are given subsets of the vertices $S_1\subset V_1$ and $S_2\subset V_2$ called \emph{seeds}  with $|S_1|=|S_2|=s$ and a bijective seeding function $\phi_S:S_1\rightarrow S_2$.  Without loss of generality we may reorder the vertices so that $S_1=S_2=[s]$ and $\phi_S=\mathrm{id}$ (the identity function on $S_1$).  The task then is to use $\phi_S$ to estimate $\phi$ by finding the bijection extending $\phi_S$ which minimizes (\ref{eq:min}).  
In the non-bijective setting, to accommodate the fact that the latent alignment need not be one--to--one, we define $\phi$ to be a subset of $V_1\times V_2$, and we are tasked with using a partial observation of $\phi$ to estimate the remaining latent alignment.

\section{Divide-and-conquer seeded graph matching}
We present the details of the LSGM algorithm, Algorithm \ref{alg:dandc}.
In section~\ref{sec:jclust}, we describe Steps 1-3 of this algorithm which constitute the divide steps.
In section~\ref{sec:matchClust}, we describe the final step of the algorithm which constitutes the conquer step.
\label{S:SEC}
\subsection{Jointly embedding and clustering the graphs} \label{sec:jclust}
We begin by describing the embedding and clustering subroutine. The input is the symmetric
 adjacency matrices $A$ and $B$ of the two graphs to be matched ($G_1$ and $G_2$ respectively), the number of seeds
 $s\in\mathbb{Z}^{+}$, the seeding function $\phi_S:[s]\rightarrow[s]$,
  the number of clusters $k$, and the embedding dimension
 $d\in\mathbb{Z}^+$. 
 Note that the procedure can easily be modified to handle directed graphs as well.
\begin{algorithm}
\caption{Jointly embedding and clustering the vertices of two graphs, $G_1$ and $G_2$}         
\label{alg:emcl}                           
\begin{algorithmic}
    \STATE{\bf INPUT:} Symmetric $A, B \in\{0,1\}^{n\times n}$, $s\in\mathbb{N}$, seeding $\phi_S:[s]\rightarrow[s]$, $d\in\mathbb{N}$, $k\in[n]$; 
    \STATE{\bf OUTPUT:} A clustering of the $2n$ embedded vertices into $k$ clusters;   
    \STATE {\bf Step 1:} Compute the first $d$ orthonormal eigenpairs of $A$ and $B$, namely $(U_A,S_A)$ and $(U_B,S_B)$ respectively; 
    \STATE {\bf Step 2:} $\hX\leftarrow U_A S_A^{1/2}$, $\hY\leftarrow U_BS_B^{1/2}$;
        \STATE  {\bf Step 3:} $ \hX_s\leftarrow \hX([s],:)$, $\hY_s\leftarrow \hY([s],:)$, $Q\leftarrow\text{argmin}_{W\in W(d)}\|\hX_sW-\hY_s\|_F$;
\STATE {\bf Step 4:} Apply the transformation $Q$ to $\hX$ obtaining the embedding $\hX Q$ of $A$;
\STATE {\bf Step 5:} Cluster the $2n$ embedded points, $\{\hX Q(i,:),\,\hY(i,:)\}_{i=1}^n$ into $k$ clusters via the $k$-means clustering procedure;
\end{algorithmic}
\end{algorithm}

\noindent{\bf Step 1:}  Compute the first $d$ eigenpairs of $A$ and $B$.  Letting the orthonormal eigen-decompositions of $A=[U_A|\widetilde U_A](S_A\oplus \widetilde S_A)[ U_A|\widetilde U_A]^T$ and $B=[U_B|\widetilde U_B](S_B\oplus \widetilde S_B)[ U_B|\widetilde U_B]^T$, with $U_A,\,U_B\in\mathbb{R}^{n\times d}$, $S_A,\,S_B\in \mathbb{R}^{d\times d}$ and the diagonals of $(S_A\oplus\widetilde S_A)$ and $(S_B\oplus \widetilde S_B)$ nonincreasing, we compute only $U_A$, $U_B$, $S_A$, $S_B$.  

\noindent{\bf Step 2:}  Initially embed the vertices of $G_1$ and $G_2$ into $\mathbb{R}^d$ as $\hX:=U_A S_A^{1/2}$ and $\hY:=U_B S_B^{1/2}$ respectively.  

\noindent{\bf Step 3:}  Let $ \hX_s:=\hX([s],:)$ and $\hY_s:=\hY([s],:)$ be the initial embedding of the seeded vertices.  Align the embedded seeded vertices via the orthogonal Procrustes fit problem: for 
$W(d):=\{W\in\mathbb{R}^{d\times d}:W^TW=I\}$, we set
$Q=\text{argmin}_{W\in W(d)}\|\hX_sW-\hY_s\|_F.$

\noindent{\bf Step 4:}  Align the two embedded adjacency matrices; i.e. we apply the transformation $Q$ to $\hX$ and obtaining the transformed embedding $\hX Q$.      

\noindent{\bf Step 5:} Cluster the $2n$ embedded vertices, $\hX Q$ and $\hY$, into $k$ clusters with the \textit{k-means} algorithm (\cite{kmeans}).  Let the corresponding cluster centroids be labeled $\{\mu_i\}_{i=1}^k$.

\bigskip

\noindent {\bf Remark 3.1.1.}
The above procedure can be implemented on very large graphs using efficient SVD algorithms (see for example \cite{svd}).  Indeed, as we are only interested in the first $d\ll n$ eigenpairs, these can be computed in $O(n^2 d)$ steps for $d\leq\sqrt{n}$.  In the sparse regime, fast partial singular value decompositions (e.g. IRLBD in \cite{ssvd3}) can be effectively implemented on arbitrarily large graphs.  Paired with fast clustering procedures (here, each iteration of $k$-means has complexity $O(dkn)$, and in practice excellent performance can often be achieved with significantly less than $n$ iterations), the above procedure can be effectively run on extremely large sparse graphs.

We do not implement parallelized versions of the SVD procedure or clustering procedure in our algorithm; indeed, even for the large graphs we considered, the partial SVD and direct $k$-means were directly and efficiently computable.  Note that there is an extensive literature devoted to parallel SVD and clustering implementations, see \cite{parsvd} and \cite{km|} for more detail.  Empirically, we see that the matching step is the most computationally intensive step of our procedure, and the runtime gains possible by parallelizing the SVD and clustering procedures are relatively small compared to the gains achieved by matching in parallel.  See Section \ref{scalability} for detail.

Additionally, the orthogonal Procrustes problem in Step 3 can be solved in $O(nd^2)$ time as it involves computing the singular value decomposition of $\hX_s^T \hY_s= USV^T\in \mathbb{R}^{d\times d}$ and setting $Q=U^T V$.

\noindent {\bf Remark 3.1.2}  Model selection, more specifically choosing $d$ and $k$, is a difficult hurdle to overcome in spectral clustering (see \cite{stp} and \cite{rohe2011spec} for instance).  One way to estimate $d$ is via automated profile likelihood procedures such as \cite{scree}.  Unfortunately, the procedure in \cite{scree} requires computation of the full spectrum, which is computationally intensive.  
In our simulation examples we assume $d$ is known, and in the real data examples, we use the ideas of \cite{chat} to estimate the embedding dimension from a partial SCREE plot.  We expect our procedure to work well as long as $d\ll\sqrt n$ (see Lemma \ref{lem:gap} for detail) which we see is the case in our simulated and real data examples.

Our procedure is insensitive to our choice of $k$ provided that 
\begin{itemize}
\item[1.] The procedure consistently clusters across the graphs---if the optimal matching of $G_1$ and $G_2$ is given by $\phi:V_1\mapsto V_2$ (in the bijective case), then for all $v\in V_1$, $v$ and $\phi(v)$ are in the same cluster.  This is essential for ensuring the accuracy of the subsequent matching step.
\item[2.] The clusters are modestly sized (for implementing the subsequent matching procedure).
\end{itemize}
Note that in practice it is impossible to ensure that the clustering is consistent, and we explore the impact of different values for $k$ (and misclustered vertices) in Section \ref{sec:k}.
Indeed, the accuracy of the algorithm is limited by the initial clustering, and we are presently working to understand the consistency of different clustering procedures in different model settings. 

\noindent {\bf Remark 3.1.3.} Practically, the particular choice of clustering procedure utilized in Step 5 of Algorithm \ref{alg:emcl} is of secondary importance.  Indeed, we choose the $k$-means clustering procedure (using {\it Matlab}'s built in k-means solver) because of its ease of implementation and theoretical tractability.  The particular clustering procedure can be chosen to optimize speed and accuracy given the properties of the underlying data.   See \cite{duda} for a review of clustering procedures.
Also note that although in many applications a natural $k$ is dictated by the data, we do not need to exactly find $k$.  For our graph matching exploitation task we do not need to finely cluster the vertices of our graphs; a gross but consistent clustering would still achieve excellent performance. 

\noindent {\bf Remark 3.1.4.} While our algorithm is presented for undirected unweighted graphs, we could adapt our approach to directed graphs (we would embed the vertices as in \cite{stfp}), or weighted graphs (the SVD can easily be run on weighted graphs).  We plan to theoretically explore this further in future work.

\subsection{Matching within clusters}\label{sec:matchClust}

When the desired matching is bijective, we first must resolve disagreements in cluster sizes and adjust the clusters accordingly.
More specifically, we need to address the fact that within each cluster, we may have an unequal number of vertices from each of the two graphs.  
We do this as follows:  
\begin{itemize}
\item[i.]
Suppose that for each $i=1,2,\ldots,k$, cluster $i$ has $c_i$ total vertices (from both graphs combined) with $c_1\geq c_2\geq\cdots\geq c_k$.  Within cluster $i$, suppose there are $c_i^{(1)}$ vertices from $G_1$ and $c_i^{(2)}$ vertices from $G_2$.  
\item[ii.]
Resize cluster $i$ to be of size 
\begin{equation}
\label{eq:blocksize}
\widetilde c_i=2\bigg\lceil\frac{c_i^{(1)}+c_i^{(2)}}{2}  \bigg\rceil-2\cdot\mathbbm{1}\bigg\{\sum_{j=1}^k \bigg\lceil\frac{c_j^{(1)}+c_j^{(2)}}{2}  \bigg\rceil\geq i+n\bigg\}.
\end{equation}  
To parse out Eq.\! (\ref{eq:blocksize}), note that
ideally we would resize the clusters to be of size $\bigg\lceil\frac{c_i^{(1)}+c_i^{(2)}}{2}  \bigg\rceil$, but $\sum_{i}\bigg\lceil\frac{c_i^{(1)}+c_i^{(2)}}{2}  \bigg\rceil$ may be greater than $n$ (note that it is never greater than $n+2k$).  To account for this, we sequentially (starting from the smallest cluster and working up) remove 2 vertices from each cluster until $\sum_i \tilde c_i=n$.  
\item [iii.] Designating all vertices as unassigned, sequentially for $i=1,2,\ldots,k$, assign the $\widetilde c_i/2$ unassigned vertices from each graph closest (in the L$^2$ sense) to $\mu_i$ to be in cluster $i$.  
\end{itemize}
Note that if the desired output is a non-bijective matching, the above procedure for ameliorating cluster sizes need not be implemented.

Once the cluster sizes are resolved, we can match the two graphs within a cluster using any number of bijective matching algorithms.  See Section \ref{S:results} for performance comparisons of various matching procedures.
These matching sub-routines can be run fully in parallel, and if the matching within cluster $i$ is denoted $\psi_i$, then the final output of our algorithm is the full matching $\psi=\oplus_{i=1}^K \psi_i$, an approximate solution to the SGMP.  To further parallelize our approach, one could implement a multithread graph matching procedure as in \cite{khan2012multithreaded}.   However, to run their procedure one needs a machine with a NUMA architecture and OpenMP installed, whereas we focus on a scalable procedure able to be run on a typical computer cluster, without any specialized hardware/software.

\noindent {\bf Remark 3.2.1}  First, note that the distances needed to resize the cluster have already been computed by the $k$-means clustering procedure so that the cost incurred by reassigning the vertices is computationally minimal (see Section \ref{S:results} for empirical evidence of this). Second, we do not focus on modifying existing $k$-means procedures to automatically make the clusters be of commensurate sizes.  We view our resizing as a refinement of the original $k$-means procedure, and not as providing a new clustering of the vertices.  In practice, our reassigned clusters are very similar to the original $k$-means clusters, often differing in only a few vertices.  

\noindent {\bf Remark 3.2.2}  In the event that one of the $k$-means clusters is composed of a large majority of vertices from a {\em single} graph,
bijective graph matching might not be sensible.  In this case, we can non-bijectively match within each cluster by padding the adjacency matrices with empty vertices to make the graphs of commensurate size (as suggested in \cite{path}), and match the resulting graphs.  Vertices matched to isolates could be treated as unmatched, or we could iteratively remove the matched vertices in the larger graph and rematch the graphs, yielding a many--to--many matching.

\noindent {\bf Remark 3.2.3} In these matching procedures, it is not surprising that we obtain best results if we use the seeded vertices to not only cluster but also match the graphs (via the SGM algorithm of \cite{FAP} and \cite{sgm2}).  We recognize that the other bijective matching procedures (\cite{path} and \cite{jovo}) have not been modified in the literature to accommodate seeded vertices, and we do not pursue the modification here.  Our results point to the need for modifying these algorithms to handle seedings, and we expect them to achieve excellent performance when thus modified. 

\subsection{Computational cost of LSGM}
\label{cost}
The many executions of the bijective matching subroutine can be run in parallel, and if $\tilde c$ is the size of the largest cluster of the points, then this step has computational complexity $O((\tilde c+s)^3)$ (assuming that we use all seeds in the matching procedure). If the executions are run in sequence then this step would have complexity $O(k(\tilde c+s)^3)$.   If $\tilde c=\Theta(n)$ then the computational cost of this step is $O(n^3)$, and we have the same computational bound as the algorithms of \cite{FAQ}, \cite{path}, \cite{jovo}.
To deal with this issue of load balancing, we re-cluster any overly large clusters by re-running our embedding and clustering procedure with the same seeding function $\phi$ on (where $\ell_i$ is the set of indices of the unseeded vertices in cluster $i$)
$$A_i=\begin{pmatrix} A'([s],[s])&A'([s],\ell_i)\\ (A'([s],\ell_i))^T&A'(\ell_i,\ell_i) \end{pmatrix},$$ 
 and $B_i$ (defined analogously) for all $i$ such that the size of the corresponding cluster is overly large.
 If we are unable to reduce these cluster sizes further, then our algorithm cannot improve upon the existing $O(n^3)$ computational complexity, though we achieve a significantly better lead constant.  In this case, we might overcome this hurdle by non-bijectively matching any overly large clusters, as these procedures are often highly scalable.

 \noindent{\bf Remark 3.3.1.} If there exists an $\alpha>0$ such that $s=o(n^{1-\alpha})$, $k=\Omega(n^{\alpha})$ and each cluster is size $O(n^{1-\alpha})$, then the computational cost of the LSGM algorithm is $O(n^2d)$ for $\alpha\leq 1/3$ and $O(n^{3(1-\alpha)})$ for $\alpha>1/3$ when the matching subroutines are fully parallelized.  Hence, a modest number of modestly sized clusters---$\alpha\approx1/3$---yields a $O(n^2d)$ running time for the LSGM algorithm. 

\subsection{Active seed selection}
If the number of seeds is large and if the seeds are all used in the matching procedures (i.e. we use SGM to match the clusters), the LSGM algorithm may be computationally unwieldy.  To remedy this, we formulate a procedure for active seed selection that aims to optimally choose a computationally tractable number of seeds from $S$ to match across each cluster.  If we are matching cluster $i$ of size $c_i$ across $G_1$ and $G_2$, and computationally we can only handle an additional $s_i$ seeds in the SGM subroutine---so that we are matching $c_i+s_i$ total vertices---then ideally we would want to pick the ``best" $s_i$ seeds to use.  Luckily, the results of $\cite{sgm2}$ provide a useful heuristic for what defines ``best" in this setting.

Ideally, columns of the seed to non-seed adjacency matrix in $G_1$ and $G_2$ would be enough to uniquely identify the unseeded vertices in each graph and this can be achieved with a logarithmic number of randomly chosen seeds \cite{sgm2}. 
Though this is a limiting result, the result (and its proof) offers insight into how to select the ``best'' seeds in a finite resource setting.  Specifically, we seek to have the columns of the seed to non-seed adjacency matrix maximally distinguish the unseeded vertices.  Mathematically, this translates to choosing seeds that have the maximum entropy in their collection of seed-nonseed adjacency vectors.  To this end, we formulate the following seed selection algorithm for selecting the seeds to use when matching across cluster $i$ (for $i$ fixed).

Suppose that the desired number of seeds for matching cluster $i$ is $s_i$.
To have the columns of the seed to non-seed adjacency matrix maximally distinguish the unseeded vertices, we seek seeds that have maximum entropy contained in their collection of seed-nonseed adjacency vectors.  
We propose to accomplish this greedily by repeatedly maximizing the (average across the two graphs) entropy increase possible by adding a {\it single} inactive seeded vertex to our active seed set.  Abusing notation, define 
\begin{equation}
\label{eq:ent}
H^j(\mathcal{S}_i)=H\bigg[A_j(\mathcal{S}_i,\ell_i^j)\bigg],
\end{equation} 
to be the Shannon entropy of the binary column vectors of the seed to nonseed adjacency matrix in graph $G_j$ with seed set $\mathcal{S}_i\subset \mathcal{S}$ and unseeded vertices $\ell_i^j$ and $H$ is the Shannon entropy function.
Initialize $\mathcal{S}_i^{(0)}=\emptyset$ and for $t=1,2,\ldots,s_i$, we set $\mathcal{S}_i^{(t)}$ to $\mathcal{S}_i^{(t-1)}\cup \{i_t\}$ where 
\begin{equation}
\label{eq:seedsel2}
i_t\in\text{argmax}_{i\in[s]\setminus \mathcal{S}_i^{(t-1)}}\left(H^1(\mathcal{S}_i^{(t-1)}\cup\{i\})+H^2(\mathcal{S}_i^{(t-1)}\cup\{i\})\right).
\end{equation}  
Finally, set $\mathcal{S}_i = \mathcal{S}_i^{(s_i)}$.

For example, suppose that we have 4 seeded vertices and 4 unseeded vertices and seed to nonseed adjacency given by:
$$A([s],C_i^1)=\begin{pmatrix} 1&0&1&0\\0&1&1&0\\1&1&1&0\\0&1&0&0\end{pmatrix}, B([s],C_i^2)=\begin{pmatrix} 1&1&1&0\\1&0&1&0\\1&1&0&0\\1&1&1&0\end{pmatrix}.$$
If we were choosing 3 seeds for subsequent matching, we would choose (in this order): $i_1=2$, then $i_2=1$ (seed 3 could also have been chosen as there are two maximizers of the entropy), then $i_3=3.$

\section{LSGM and the Stochastic Block Model}
\label{s:sbm}
In as much as we can partition the vertices of $G_1$ and $G_2$ into consistent clusters, it is natural to model $G_1$ and $G_2$ using the \emph{stochastic block model} (SBM) of \cite{sbm} and \cite{sbm2} (details of the model are presented shortly).  We then define the clustering criterion for clustering the rows of $[\hY^T|(\hX Q)^T]^T$ into $k$ clusters via
\begin{align}
\label{eq:cc}
(\widehat C,\hat b):=\text{argmin}_{C\in\mathbb{R}^{k\times d},\ b:\,[2n]\rightarrow[k]}\sum_{i=1}^{2n}\bigg\|\left[ \begin{pmatrix} 
  \hY \\
  \hX Q  
\end{pmatrix} \right](i,:)-C(b(i),:)\bigg\|^2_2,
\end{align}
where the rows of $\widehat C$ are the centroids of the $k$ clusters and $\hat b$ is the cluster assignment function. Note that $k$-means attempts to solve (\ref{eq:cc}).  In Theorem \ref{T} we show that, under some mild conditions on the underlying SBM, the optimal cluster assignment $\hat b$ almost surely perfectly clusters the vertices of both $G_1$ and $G_2$.
We present the necessary background below.

A $d$-dimensional stochastic block model random graph, $G$, has the following parameters:
an integer $K \geq 2$, a vector of nonnegative 
integers $\vec{n} \in \mathbb{N}^K$, and a latent--position matrix $X\in[0,1]^{n\times d}$ with $K$ distinct rows.
The random graph's vertex set $V$ is the union of 
the {\it blocks} $V_1$, $V_2$, \ldots, $V_K$, which are  
disjoint sets with respective cardinalities 
$n_1$, $n_2$, \ldots, $n_K$. For each $v \in V$, let $b(v)$ denote the 
block of $v$, ie $v \in V_{b(v)}$. Lastly, for each 
 pair of vertices $\{v,v' \} \in {V \choose 2}$, the 
adjacency of $v$ and $v'$ is an independent Bernoulli trial 
with probability of success $\gB(v,v')$, where $\gB:=XX^T$.  

Two independent SBM graphs may have no correlation structure between them, and there is no natural bijective alignment of their vertices.  To induce this alignment, we introduce correlation between the graphs.  We say that two (matched) random graphs $G_1$ and $G_2$ from this model
have correlation $\rho \in [0,1]$ if the set of indicator random variables 
$$\{\mathbbm{1}_{v \sim_{G_1} v'},\mathbbm{1}_{w \sim_{G_2} w'}\}_{\{v,v'\},\{w,w'\}\in\binom{V}{2}}$$
are mutually independent except that for each  
$\{v,v' \} \in {V \choose 2}$, the indicator random variables 
$\mathbbm{1}{v \sim_{G_1} v'}$ and $\mathbbm{1}_{v \sim_{G_2} v'}$
have Pearson product-moment correlation coefficient $\rho$.
Such correlated graphs can be easily constructed by realizing $G_1$
from the underlying SBM and then, for each $\{v,v' \} \in {V \choose 2}$,
$\mathbbm{1}_{v\sim_{G_2} v'}$ is an independent 
Bernoulli trial with probability of success 
$\gB(v,v')+\rho (1-\gB(v,v'))$ if 
$v$ and $v'$ are adjacent in $G_1$, and probability of success
$\gB(v,v')(1-\rho)$ if $v$ and $v'$ are not 
adjacent in $G_1$.  If $G_1$ and $G_2$ are thus correlated, then there is a natural latent alignment between the vertices of the graphs, namely the identity function id$_n$.

Given $\vec{m} \in \mathbb{N}^K$ such that 
$\vec{m} \leq \vec{n}$ coordinate-wise and $\|\vec{m}\|_1=s$ (the number of seeds), the random graphs 
$G_1$ and $G_2$ from the $d$-dimensional stochastic block model parameterized 
with $K$, $\vec{n}$, $X$, and having correlation $\rho$,
are $\vec{m}$-seeded if, a priori for each $i=1,2,\ldots,K$,
\  $m_i$ of the $n_i$ vertices from block $V_i$ function as seeds for LSGM, i.e.~their across graph correspondence is known.

Let $G_1$ and $G_2$ be $\rho$-correlated, $\vec m$-seeded (with $\vec m^T\vec{1}=s$), $d$-dimensional SBM's parametrized by $K$, $\vec n$, and $X$.   Let their respective adjacency matrices be $A$ and $B$, and let their respective block membership functions be $b_A$ and $b_B$.  Without loss of generality, let the true alignment function be id$_n$ and let $b:=b_A=b_B$.   Consider the transformed (as in Step 4 of Algorithm \ref{alg:emcl}) adjacency spectral embeddings of $G_1$ and $G_2$, $\hX Q$ and $\hY$, and assume that we have clustered the rows of $[\hY^T|(\hX Q)^T]^t$ via the optimal $(\widehat C,\hat b)$ of (\ref{eq:cc}).  Adopting the notation of Algorithm \ref{alg:dandc}, define (where again $C_i^j$ is the set of unseeded indices in $G_j$ corresponding to cluster $i$ and $c_i=|C_i^j|$)
\begin{align}
\label{eq:optclust}
\psi^{(i)}_s &:=\text{argmin}_{P\in\Pi(s+c_i)} \big\|\left(\begin{smallmatrix} A_s&A([s],C_i^1)\\A([s],C_i^1)^T& A^{(i)}\end{smallmatrix}\right)-\left(\begin{smallmatrix} I_s&0\\0& P\end{smallmatrix}\right)\left(\begin{smallmatrix} B_s&B([s],C_i^2)\\B([s],C_i^2)^T& B^{(i)}\end{smallmatrix}\right)\left(\begin{smallmatrix} I_s&0\\0& P^T\end{smallmatrix}\right)\big\|_F,\\
\psi^{(i)}_n &:=\text{argmin}_{P\in\Pi(c_i)} \big\|A^{(i)}-PB^{(i)}P^T\big\|_F
\end{align}
to be the respective optimal seeded and unseeded matchings of cluster $i$ across the two graphs.  When appropriate, we will drop the subscript and refer to the matching of cluster $i$ as simply $\psi^{(i)}$.

We shall hereto forth be considering a sequence of growing models with $n=1,2,\ldots$ vertices.  In the next theorem, we prove that under modest assumptions, we have that for all but finitely many $n$, $\hat b=b$, and all of the vertices are perfectly clustered across the two graphs.  The results of \cite{sgm2} immediately give that $\psi^{(i)}_s=\{I_{s+c_i}\}$ a.a.s.\@ and $\psi^{(i)}_n=\{I_{c_i}\}$ a.a.s.\@ for all $i=1,2,\ldots, K$ and the above procedure (when perfected implemented) correctly aligns the two SBM graphs.Although this result is asymptotic in nature, it provides hope that our two-step procedure will be effective in approximating the the true but unknown alignment across a broad spectrum of graphs.  

\begin{theorem}
\label{T}
With notation as above, let $G_1$ and $G_2$ be $\vec m$-seeded (with $\vec m^T\vec{1}=s$), $d$-dimensional SBM's parametrized by $K$, $\vec n$, and $X$.  Although we assume $G_1$ and $G_2$ have the same block structure, we make no assumptions about the correlation structure.  Let their respective adjacency matrices be $A$ and $B$, and without loss of generality let the true alignment function be $id_n$, so that the  block membership function is $b:=b_A=b_B$.  Adopting the notation of Section \ref{S:SEC}, if the following assumptions hold:
\begin{itemize}
\item[i.] There exist constants $\epsilon_1,\, \epsilon_2>0$ such that $K=O(n^{1/3-\epsilon_1})$ and $\min_i\vec{n}(i)=\Omega(n^{2/3+\epsilon_2})$;
\item[ii.] Defining 
\begin{equation}
\label{eq:gap}
\delta_d:= \min_{i,j\leq d+1, i\neq j}|\lambda_i(XX^T)-\lambda_j(XX^T)|/n,
\end{equation} 
and
\begin{equation}\label{eq:beta}\beta:=\beta(n,d, \delta_d)= \frac{260 d\log(n)}{\delta_d n^{1/2}}, \end{equation}
if $i,\,j\in[n]$ are such that $X(i,:)\neq X(j,:)$ then $\|X(i,:)-X(j,:)\|_2>6n^{1/6} \beta$;
\item[iii.] Without loss of generality, let $\{X(i,:)\}_{i=1}^s$ be the latent positions corresponding to the seeded vertices, then we assume there exists an $\alpha$ satisfying $\alpha>4\beta$ and $\sqrt{n}\beta/\alpha=o(n^{\epsilon_2/2}d/\delta_d)$ such that 
\begin{equation}
\label{eq:latsep}
\min_{v\, :\, \|v\|_2=1}\|X([s],:)v^T\|_2\geq \alpha\sqrt{s};
\end{equation}
\end{itemize}
\noindent then for all but finitely many $n$, 
the $\hat b$ of (\ref{eq:cc}) satisfies $\hat b=b.$ 
\end{theorem}
\begin{figure}[t!]
\hspace{0mm}
\includegraphics[width=1\textwidth]{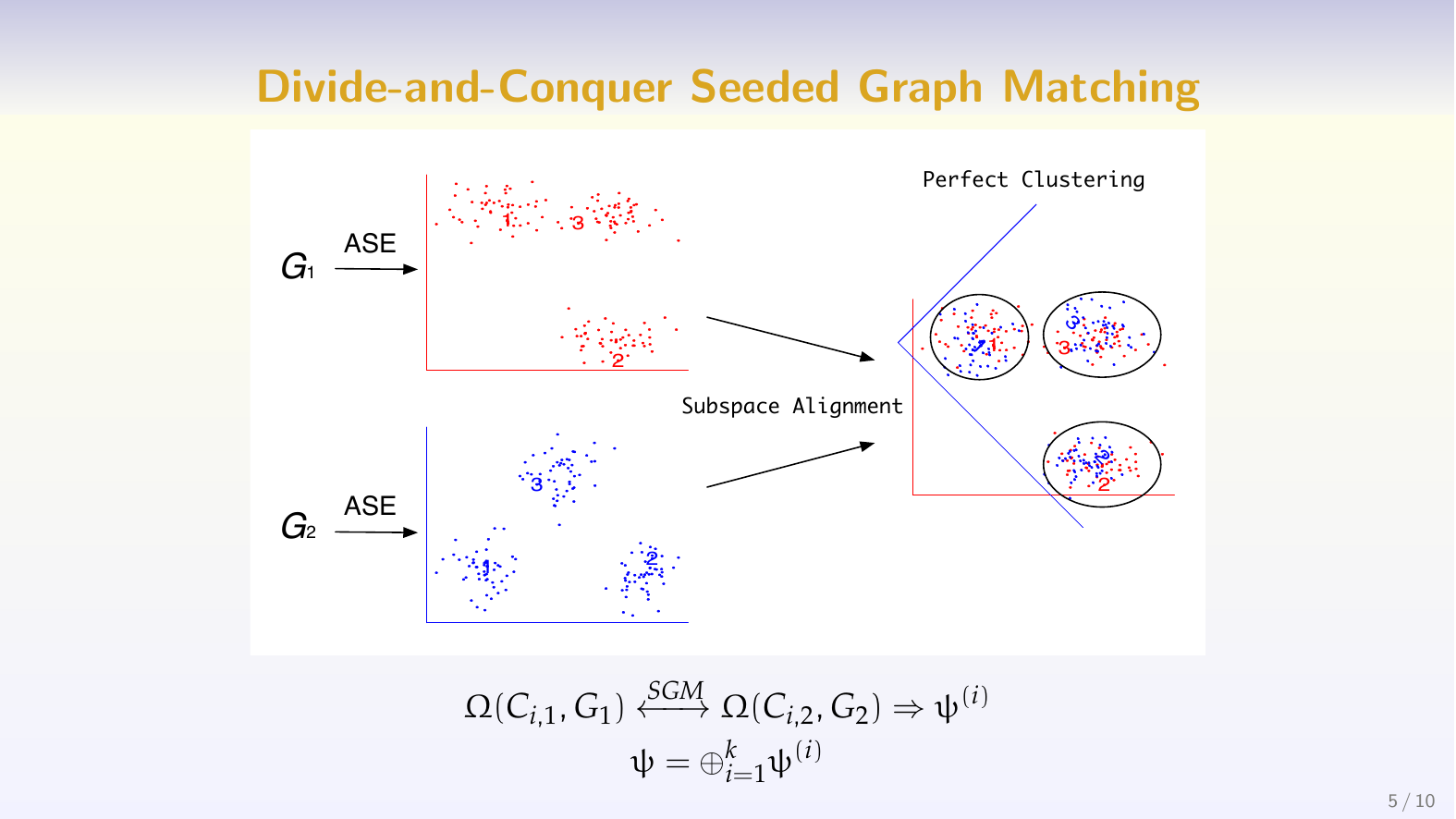}
\caption{Visual proof sketch of Theorem \ref{T}.  The graphs are first embedded using Adjacency Spectral Embedding (ASE), aligned using the seeded vertices, and perfectly clustered in the aligned space.}
\label{fig:sgmlsgm}
\end{figure}
 Regardless of the correlation structure, Theorem \ref{T} implies that our joint clustering procedure yields a canonical nonbijective matching of the vertices (where the matching is given by the clustering).  

Our proof of this theorem will proceed as follows.
First we will state some key results proved elsewhere. 
Then we will bound  $\|\widehat{X}Q-\widehat{Y}\|_{2\to \infty} := \max_{i} \| (\widehat{X}Q-\widehat{Y})_{i\cdot}\|_2$ and
will then have that the $2n\times d$ matrix $[\widehat{Y}^T | (\widehat{X}Q)^T]^T$ is close to a specified transformation of the $[X^T |X^T ]^T$  (recalling from \cite{perfectclust} that for a matrix $M\in\mathbb{R}^{a\times b}$, 
$\|M\|_{2\rightarrow\infty}=\max_{i}\| M(i,:)\|_2$).
Finally, we will use this to show that the clustering will perfectly cluster the vertices in the two graphs into the $K$ true blocks.

Let $\gB=[U_{\gB}|\widetilde U_{\gB}][S_{\gB}\oplus\widetilde S_{\gB}][U_{\gB}|\widetilde U_{\gB}]^T$ 
be the orthonormal eigen-decomposition of ${\gB}$ with
 $U_{\gB}\in\mathbb{R}^{n\times d},$ $S_{\gB}\in\mathbb{R}^{d\times d}$,
  and ordered so that the diagonals of $[S_{\gB}\oplus\widetilde S_{\gB}]$ are nondecreasing. 
 The next lemma collects some necessary results from \cite{stfp} and \cite{perfectclust} which will be needed in the sequel.  

\begin{lemma}
\label{lem:gap}
With notation as above, 
let $W_A = \argmin_{W\in W(d)} \|\hX-XW\|_F$ and \\$W_B=\argmin_{W\in W(d)} \|\hY-XW\|_F$.
If $d=o(\sqrt n),$ then it holds with probability one that for all but finitely many $n$ that
\begin{align}
\label{eq:l9}
\|\hX-X W_A\|_{2\rightarrow \infty}\leq \beta \text{ and }
\|\hY-X W_B\|_{2\rightarrow \infty}\leq \beta.
\end{align}

\end{lemma}

We are now ready to prove the following.
\begin{lemma}\label{lem:XQ-Y}
For all but finitely many $n$ it holds that
$\|\widehat{X}Q-\widehat{Y}\|_{2\to\infty}\leq  8\beta/\alpha+2\beta.$
\end{lemma}
\noindent{\it Proof:}
As in Section \ref{S:SEC}, let $Q:=\argmin_{W\in W(d)}\|\hX([s],:)W-\hY([s],:)\|_F$
and let $\tilde{Q} = W_A^\top W_B$.
It immediately follows from Eq.~\eqref{eq:l9} that $\| \hX \tilde{Q}- \widehat{Y}\|_{2\to \infty}\leq2\beta$.
Clearly 
\begin{align}
\label{eq:boundr}
&\|\hX([s],:)Q-\hY([s],:)\|_F\leq\|\hX([s],:)\tilde{Q}-\hY([s],:)\|_F\leq 2\beta \sqrt{s}.
\end{align}
and working in the other direction
\begin{align}
\label{eq:boundl}
2\beta\sqrt{s}\geq \|\hX([s],:)Q-\hY([s],:)\|_F 
&\geq \|\hX([s],:)(Q-\tilde{Q})\|_F-\|\hX([s],:)\tilde{Q}-\hY([s],:)\|_F \notag \\
&\geq \|\hX([s],:)(Q-\tilde{Q})\|_F-2\beta \sqrt{s}.
\end{align}
If we let the SVD of $Q-\tilde{Q}$ be $V_1 S V_2^\top$ then 
\begin{align}
\label{eq:svdlb}
\|\hX([s],:)(Q-\tilde{Q})\|_F 
&\geq  \|X([s],:)W_A (Q-\tilde{Q})\|_F  -\|\left(\hX([s],:)-X([s],:) W_A\right) (Q-\tilde{Q})\|_F \notag\\
&\geq  \left(  \sum_{i=1}^s \sum_{j=1}^d \langle X(i,:),W_A V_1(:,j) \rangle S(j,j)^2  \right)^{1/2} -2\beta \sqrt{s}\|Q-\tilde{Q}\|_F \notag\\
&\geq  (\alpha-2\beta)\sqrt{s}\|Q-\tilde{Q}\|_{F}
\end{align}
by the assumption (Eq.\@ \ref{eq:latsep}) that $\min_{\|v\|_2=1} \|X([s],:) v\|_2^2\geq \alpha^2 s$ and Eq.\@ (\ref{eq:boundl}).
Combined with Eq.\@ (\ref{eq:boundr}), we have $\|Q-\tilde{Q}\|_{2\to 2} \leq \|Q-\tilde{Q}\|_F \leq \frac{4\beta}{\alpha-2\beta}$.
Hence, we have that 
\begin{align}
\label{eq:lemmaproved}
 \|\widehat{X}Q - \widehat{Y} \|_{2\to \infty}
 & \leq \|\widehat{X}(Q-\tilde{Q})\|_{2\to\infty} +\|\widehat{X}\tilde{Q} - \widehat{Y}\|_{2\to \infty}\notag\\
 &\leq \|\widehat{X}\|_{2\to \infty} 4\beta / (\alpha  - 2\beta) + 2\beta \leq 8\beta/\alpha+2\beta.\end{align}
since $\|\widehat{X}\|_{2\to \infty} \leq 1$ and $\alpha>4\beta$. \qed

\begin{lemma}\label{lem:XQY-XT}
For all but finitely many $n$, it holds that
\[ \left\|\begin{pmatrix}
\widehat{Y} \\ \widehat{X}Q
\end{pmatrix}-\begin{pmatrix}
X W_B \\ X W_B
\end{pmatrix}\right\|_{2\to\infty}\leq  \frac{8\beta}{\alpha}+3\beta. \]
\end{lemma}
\noindent{\it Proof:}
We have 
\begin{align}
\label{eq:4.4.1}
\left\|\begin{pmatrix}
\widehat{Y} \\ \widehat{X}Q
\end{pmatrix}-\begin{pmatrix}
X W_B \\ X W_B
\end{pmatrix}\right\|_{2\to\infty}
= \max \{ \|\widehat{Y}-X W_B\|_{2\to\infty}, \|\widehat{X}Q-X W_B\|_{2\to\infty} \}.
\end{align}
The first term in Eq.\@ (\ref{eq:4.4.1}) is bounded by $\beta$ by Eq.~\eqref{eq:l9}.
For the second term we have from Eq.\@ (\ref{eq:lemmaproved}) that
$
 \|\widehat{X}Q-X W_B \|_{2\to\infty} 
 \leq  \|\widehat{X}Q-\widehat{Y}\|_{2\to \infty} +\|\widehat{Y} -X W_B \|_{2\to\infty}\leq \frac{8\beta}{\alpha} + 3\beta.$\qed

\noindent{\it Pf of Main thm:}
Let $\mathcal{B}_1,\mathcal{B}_2,\ldots,\mathcal{B}_K$ be the $L^2$-balls of radius $r:=n^{1/6}\beta$ around the $K$ distinct rows of $X  W_B$. If $X(i,:)\neq X(j,:)$,
 then by assumption
\begin{align}
\label{eq:balls}
6n^{1/6}\beta&\leq \|X(i,:)-X(j,:)\|_2 = \|(X(i,:)-X(j,:)) W_B\|_2,
\end{align}
 and the $\mathcal{B}_i's$ are disjoint.  

Let $\widehat{Z} = [\widehat{Y}^T |(\widehat{X}Q)^T] ^T$ and let $Z = [(X W_B)^T|(X W_B)^T]$. 
Let $(\widehat{\mathcal{C}}, \hat b)$ be the optimal clustering of the rows of $\widehat{Z}$ from (\ref{eq:cc}).
Suppose there is an index $i\in[2n]$ such that 
$\|X(i,:) W_B-\widehat C(\hat b(i),:)\|>2r.$ 
This would imply that $\|\widehat{Z}-\widehat{C}\circ\hat b\|_F>\sqrt{\min_{j}\vec n(j)} (2r-\beta)$ (where $\widehat{C}\circ\hat b$ is the $2n\times d$ matrix whose $i-th$ row is $\widehat{C}(\hat b(i),:)$ ).  As $\min_j \vec{n}(j)=\Omega(n^{2/3+\epsilon_2})$ for a constant $\epsilon_2>0$, we would then have that
\begin{equation}
\label{eq:contr}
\|\widehat{Z}-\widehat{C} \circ\hat b\|_F=\Omega\left( \frac{n^{\epsilon_2/2}d}{\delta_d}\right).
\end{equation}
Lemma \ref{lem:XQY-XT} yields that
\begin{align}
\|\widehat{Z}-Z\|_F\leq \sqrt{2n}\left( \frac{8\beta}{\alpha} + 3\beta  \right)=o\left( \frac{n^{\epsilon_2/2}d}{\delta_d}\right),
\end{align} (where the final equality follows from assumption $iii)$.  Combined with Eq. (\ref{eq:contr}), this contradicts the minimality of $(\widehat C,\hat b),$ and therefore $\|Z-\widehat C \circ\hat b\|_{2\rightarrow\infty}\leq2r$. 

From (\ref{eq:l9}) we have 
$\|\widehat{Z}-\widehat{C} \circ\hat b\|_{2\rightarrow\infty}\leq 2r+\beta=(2+o(1))r.
$
If $i,j\in[n]$ are such that $\widehat C(\hat b(i),:)\neq \widehat C(\hat b(j),:)$, then 
$\|Z(i,:)-Z(j,:)\|_2>6r,$
and it follows that 
\begin{align}\|\widehat{Z}(i,:)-\widehat C(j,:)\|_2>4r-\beta=(4+o(1))r.
\end{align}
It follows that for all but finitely many $n$, $\hat b=[b^T|b^T]^T$.  Stated simply, 
\begin{align}\min_{\pi\in S_K} |\{v\in V(G_1)\cup V(G_2):b_n(v)\neq \pi(\hat b_n(v))\}|=0.\end{align}
Now \cite[Theorem 1]{sgm2} immediate implies that for all but finitely many $n$, $\psi^{(i)}=\{I_{u_i}\}$ for all $i\in[K]$ and the proof is complete.
\qed

 \noindent {\bf Remark 4.5.} The implication of assumption {\it iii.}\@ in Theorem \ref{T} is that in order for the scaled Procrustes fit of the embedded seeded vectors to align the entire embedding, it is sufficient that the latent positions corresponding to the seeded vectors cannot concentrate too heavily in one direction. We note that analogous assumptions are made in the literature on sparse subspace clustering, see \cite{ssc} for example and detail.

 \noindent {\bf Remark 4.6.}  If there exist  constants $\epsilon_1,\, \epsilon_2>0$ such that $K=O(n^{1/3-\epsilon_1})$ and $\min_i\vec{n}(i)=\Omega(n^{2/3+\epsilon_2})$, then the results of \cite{perfectclust} demonstrate that the optimal clustering for the one graph analogue of (\ref{eq:cc}) perfectly clusters the vertices of a single SBM. 

\section{Empirical Results}
\label{S:results}
We next explore the effectiveness of our divide-and-conquer approach on simulated and real data examples.  When comparing across graph matching algorithms, we measure effectiveness via the matching accuracy (since we assume a true latent alignment, this amounts to the fraction of vertices which were correctly aligned) and runtime of the algorithms.
Across both runtime and accuracy, our algorithm achieves excellent performance:  achieving significantly better accuracy than existing scalable bijective matching algorithms (Umeyama's spectral approach \cite{umeyama}), and achieving significantly better accuracy and runtime than the existing state-of-the-art (in terms of accuracy) matching procedures (PATH \cite{path}, GLAG \cite{jovo}, FAQ \cite{FAQ}).
Unless otherwise specified, all of our experiments are run on a 2 x Intel(R) Xeon(R) CPU E5-2660 0 $@$ 2.20GHz
(with 32 virtual cores and 16 physical cores).
We implement all of our code in the software package Matlab limited to 12 parallel threads.
Additionally, the code needed to run our algorithm (in Matlab) is publically available for download at 
\texttt{https://github.com/lichen11/LSGMcode}.
\subsection{Simulation Results}
\label{S:4}
\begin{figure}[t!]    \begin{subfigure}{.43\textwidth}
            \includegraphics[width=\textwidth]{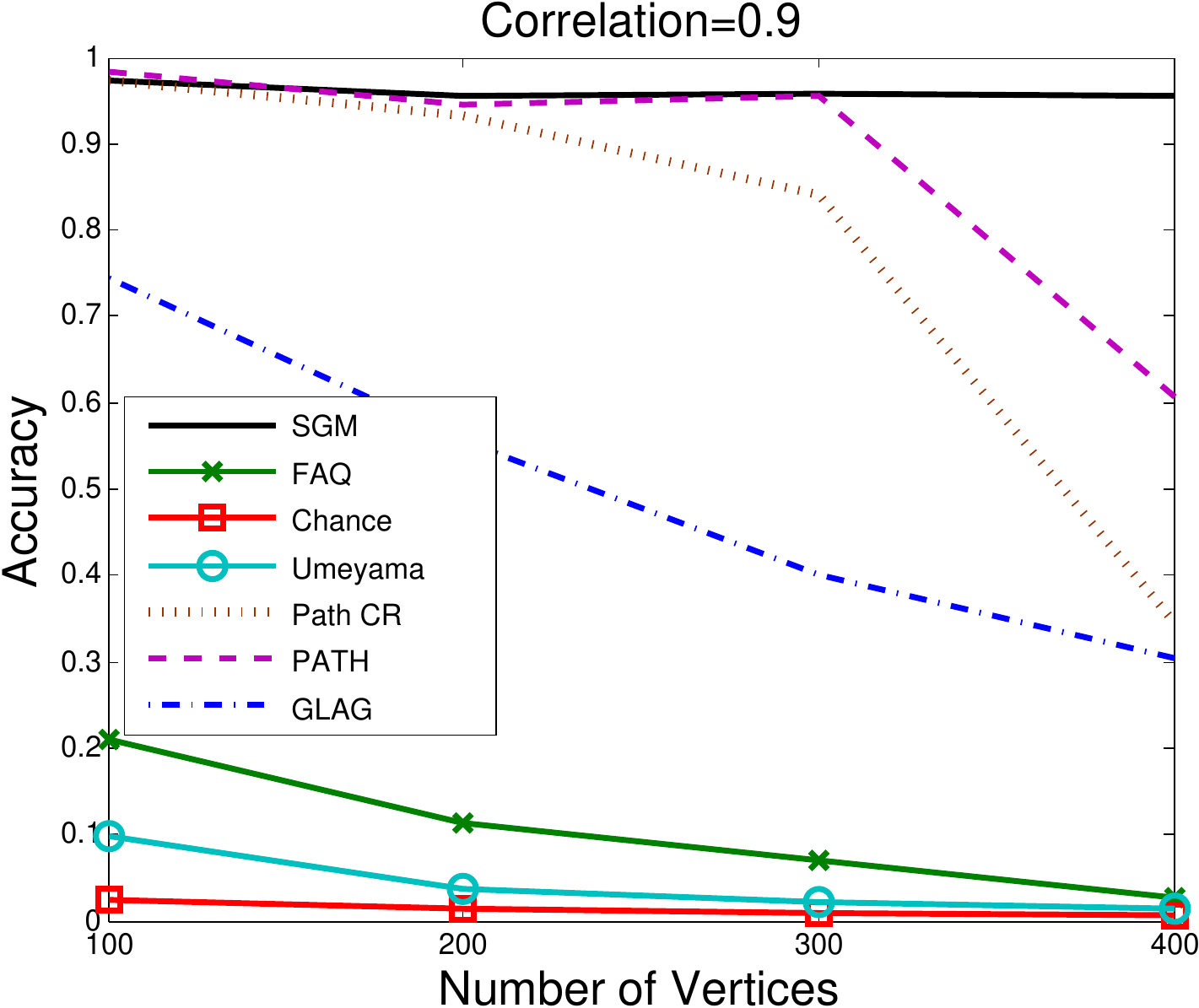}

    \caption*{Mean Running Time}
    {\small 
\begin{tabular}{r|cccc}\hline
$n=$ & 100  & 200   & 300    & 400 \\\hline
SGM                 & 0.14 & 0.78  & 2.13   & 4.49 \\
FAQ                 & 0.51 & 3.12  & 9.13   & 16.67 \\
Umeyama             & 0.09 & 0.14  & 0.21   & 0.34 \\
PATH CR             & 0.30 & 1.24  & 2.96   & 3.46 \\
PATH                & 2.21 & 9.90  & 15.82  & 69.31 \\
GLAG                & 8.53 & 33.83 & 109.48 & 261.72\\ \hline
    \end{tabular}
    }
    \end{subfigure}
    \hspace{4mm}
        \begin{subfigure}{.43\textwidth}
            \includegraphics[width=\textwidth]{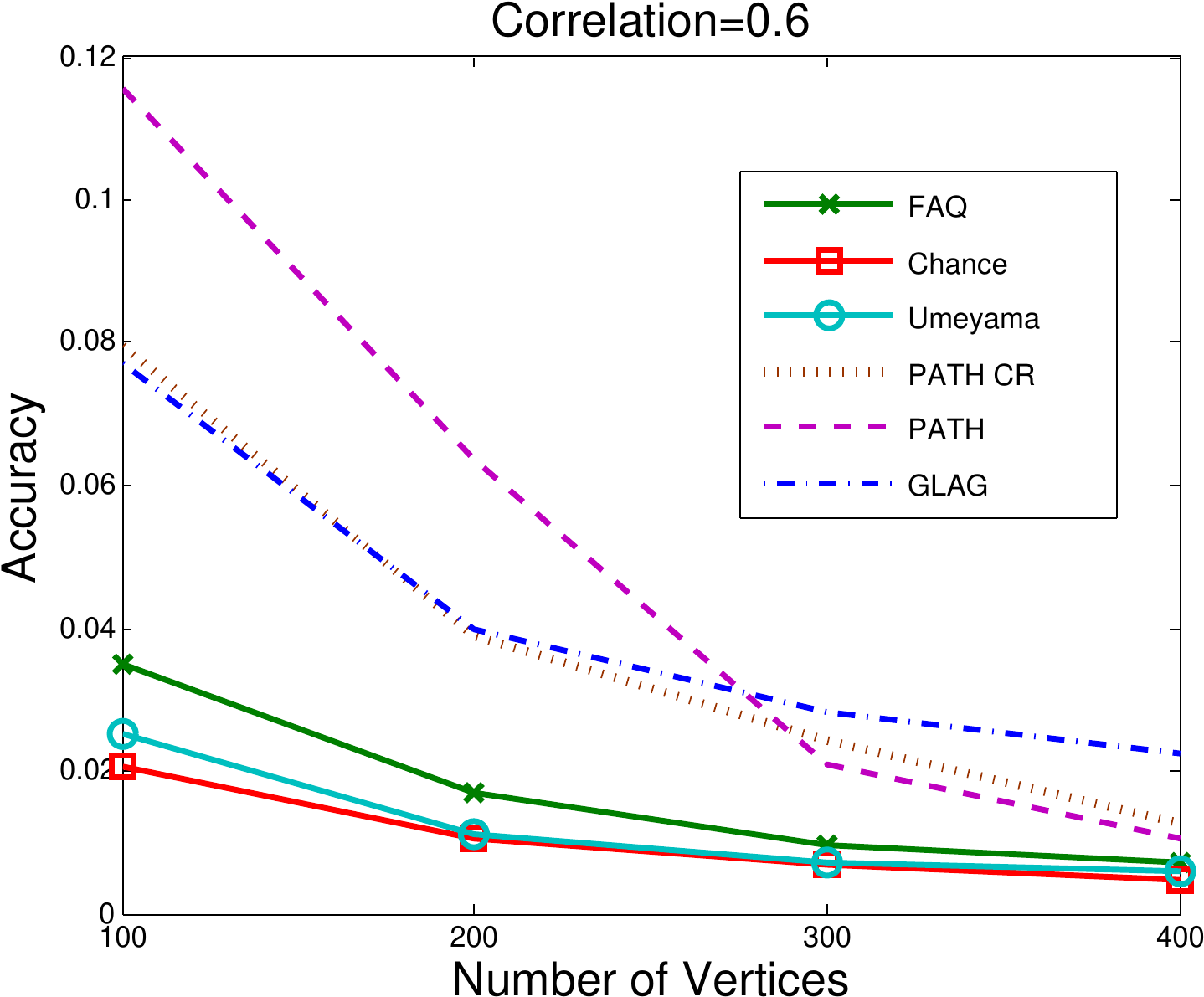}

    \caption*{Mean Running Time}
    {\small
\begin{tabular}{r|cccc}\hline
$n=$ & 100  & 200   & 300    & 400 \\\hline
SGM                 & 0.16 & 0.64  & 1.76   & 2.91 \\
FAQ                 & 0.58 & 3.05  & 9.09   & 15.56 \\
Umeyama             & 0.13 & 0.19  & 0.27   & 0.41 \\
PATH CR             & 0.42 & 1.25  & 2.73   & 2.36 \\
PATH                & 3.70 & 27.00 & 83.60  & 142.63 \\
GLAG                & 8.11 & 32.41 & 108.22 & 235.34 \\\hline
\end{tabular}
    }
    \end{subfigure}
    \caption{Mean accuracy (top) and mean runtime (bottom) for graph matching algorithms for $\rho=0.9$ (left) and $\rho=0.6$ (right). The parameters for the SBM graph are $D= I_2\otimes .3 J_{n/2}+.3 J_n$, $\vec{n}=[n/2, n/2]$, for each of $n=100,200,300,400$ and $\vec{m}=[3,3]$. For each value of $n,$ we ran 100 Monte Carlo replicates.
    Note, the difference in scales for the left and right accuracy plots. We do not include the accuracy results for SGM for $\rho=0.6$ because they are near 1 and obscure the ordering for the remaining vertices.}
    \label{fig:rho}
\end{figure}

Once the vertices of the two graphs are clustered, we can run the matching procedures in full parallel across the clusters.  
Our first experiment seeks to understand how available bijective matching algorithms perform (with respect to accuracy and speed), so that we can better understand how to appropriately set the maximum allowed cluster size.  
To this end, we run the following experiment.  We consider two $\rho$-correlated SBM random graphs with the following parameters (where $J_n:=\vec{1}_n \vec{1}_n^T\in\mathbb{R}^{n\times n}$, $I_n$ is the $n\times n$ identity matrix, and $\otimes$ denotes the Kronecker product): each of
$\rho=0.6$ and $0.9$, $D= I_2\otimes .3 J_{n/2}+.3 J_n\in \mathbb{R}^{n\times n}$, $\vec{n}=[n/2, n/2]$, for each of $n=100,200,300,400$.
We cluster the graphs into 2 clusters and run a variety of {\it bijective} GM algorithms on these clusters.  We record both the performance of the algorithms in recovering the true alignment and the corresponding running time of each algorithm.  Note we ran the matching procedures on the two clusters in parallel.
The algorithms we ran include SGM \cite{FAP}, FAQ \cite{FAQ}, the spectral matching algorithm of Umeyama \cite{umeyama}, the PATH algorithm and the associated convex relaxation (PATH CR, which is solved exactly using Frank-Wolfe methodology \cite{fw}) \cite{path}, and the GLAG algorithm \cite{jovo}).
See Figure~\ref{fig:rho} for the results.

To run LSGM, we used $\vec{m}=[3,3]$ seeds for $\rho=0.9$ and $\vec{m}=[5,5]$ seeds for $\rho=0.6$, all seeds chosen uniformly at random from the two blocks.  The seeds are always used in the embedding and clustering procedure, but SGM is the only algorithm to use seeded vertices when matching the clusters.  It is not surprising that it achieves best performance.  We expect similarly excellent results from the other matching algorithms once they are seeded.

In the $\rho=0.9$ experiment, we note that, of the nonseeded matching algorithms, PATH and its associated convex relaxation achieve the best results.  
The PATH CR procedure scales very well in running time but performs progressively worse as $n$ increases.  
On the other hand, the PATH algorithm's running time scales poorly (as does that of the
 GLAG algorithm), needing significantly longer running time than SGM or PATH CR across all values of $n$.  
While PATH and PATH CR achieve similar results to SGM for $n=100,200,300$, the significantly longer run time for PATH and the sharply decreased performance for PATH CR at $n=400$ hinder these algorithms effectiveness as post-clustering matching procedures.  
Indeed, to employ these two procedures, we would need to severely restrict the maximum allowed size of our clusters to achieve a feasible running time and/or accurate matchings.  We note that seeding GLAG, the PATH algorithm and PATH CR may yield significantly faster running times and less performance degradation as $n$ increases, as seeding FAQ yields both.

SGM is remarkably stable, achieving excellent matching performance across all $n$.  This not only indicates that our clustering methodology is consistent across graphs, but points to the importance of using the seeds in the subsequent matching.  
Here the correlation is very high, and for smaller $n,$ PATH and PATH CR perform on par with SGM, suggesting that seeds are less important when matching very similar graphs.  We next explore the effect of decreased correlation.




We explore this in the $\rho=0.6$ experiment, and again we note that SGM significantly outperforms all the nonseeded matching algorithms (with average accuracy $>99\%$ for all $n$).  This points to the consistency of our clustering procedure here.  Note that we needed slightly more seeds to achieve this consistency with the lower correlation.  Indeed, with three seeds from each cluster, the clustering was not consistent when $\rho=0.6$, unlike in the $\rho=0.9$ case.  

\subsection{Robustness to misspecified $k$}
\label{sec:k}
How sensitive is the performance of our algorithm to mis-specifying $k$?  We claim that as long as the clusters are consistently estimated, the procedure is relatively insensitive to mis-estimating $k$. 
Following this reasoning, if our clustering step allows clusters that are larger than $\max_i n_i$, then we would expect our clusters to be consistent and our performance would not degrade significantly.  However, if our clustering step does not allow cluster larger than $\max_i n_i$, then we would not expect our clusters to be consistent and our performance would degrade significantly.

\begin{figure}[t!]
\includegraphics[width=\textwidth]{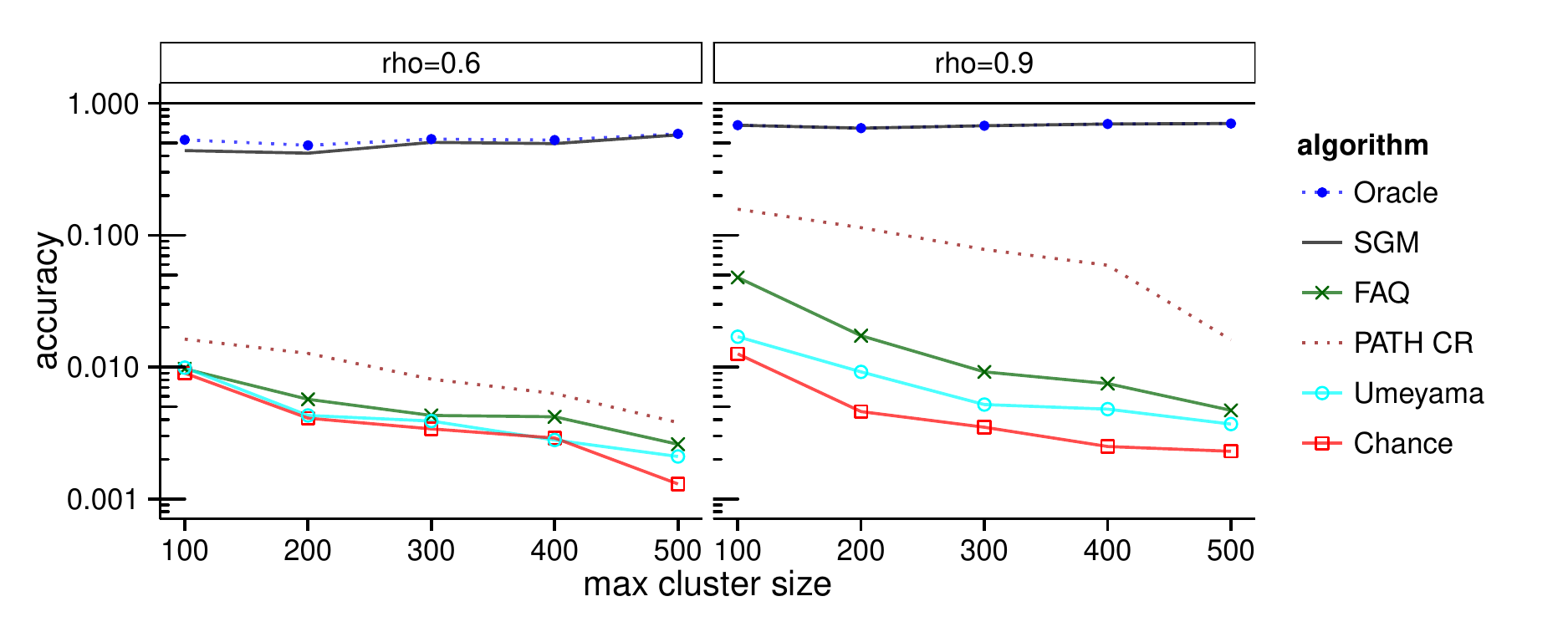}
\caption{Mean matching accuracy (on the log scale) versus maximum allowed cluster size for graph matching algorithms for $\rho=0.6$ and $\rho=0.9$, $D=I_{10}\otimes.3J_{100}+.3J_{100},$ $\vec{n}=100*\vec{1},$ and $20$ seeds randomly selected from the 1000 vertices.  For each combination of parameters, we run 20 MC replicates. Note that the oracle and SGM matching overlap heavily.  Due to scalability issues, GLAG and PATH were not run in this experiment.}
\label{fig:accVSmaxClust}
\end{figure}

To this end, we consider the following experiment.  We consider $\rho\in\{0.6,0.9\}$-correlated SBM's, with 10 blocks each of size $n_i=100$, and interblock edge probability $0.6$ and across block edge probability $0.3$.  We run 20 MC replicates of divide-and-conquer graph matching with $20$ seeds and with the maximum allowed cluster size equal to 100, 200, 300, 400, 500.  We summarize results in Figure~\ref{fig:accVSmaxClust}.  Note that we have included the ``Oracle'' matcher, which gives the maximum number of vertices possibly matched correctly given the clustering.

From the Figure~\ref{fig:accVSmaxClust}, we see that the performance of SGM again is significantly better than all the other GM algorithms considered, and is also resilient to allowing larger clusters in the $k$-means procedure.  
This is echoed in the experiment for $\rho=0.9$, where we see that SGM nearly achieves oracle accuracy across all maximum cluster sizes. 
We also explore the sensitivity of the LSGM's runtime to the maximum allowed cluster size.
Utilizing 12 cores, the average runtimes of the LSGM algorithm (using SGM for matching and $\rho=0.6$)  are
$(10.2831,   24.0464,   41.2820,   61.8609,   86.1164)$ seconds for max cluster size equal to $(100, 200, 300, 400, 500)$; indeed,
SGM has runtime $O(n^3)$ and is the slowest step of our divide-and-conquer procedure, so we expect to see the runtime increase if the matching subroutines are between bigger graphs.  
Larger clusters may be more consistent and therefore may lead to better matching performance, but this is achieved at the expense of increased runtime.

\subsection{LSGM vs. SGM: The price of embedding}
While each of PATH, PATH CR, FAQ and GLAG perform significantly better than chance, again PATH and GLAG scale poorly in running time.  The PATH CR algorithm and FAQ scale well in running time but have their matching performance decrease significantly as $n$ increases.  PATH and GLAG also see this performance degradation in $n$.  In addition, all the algorithms (except SGM) perform significantly worse than the $\rho=0.9$ case.  As real data is, at best, weakly correlated, this points to the primacy of seeding in matching real data graphs. 
Due to the decreased performance and poor scalability of the nonseeded matching algorithms as $n$ increases, we will henceforth focus our attention on using SGM to match the clusters.  Again, we expect the best performing unseeded algorithms (PATH, PATH CR and GLAG) will achieve excellent performance when seeded, though we do not pursue this modification here.  

\begin{figure}[t!]
\hspace{35mm}
\includegraphics[width=.5\textwidth]{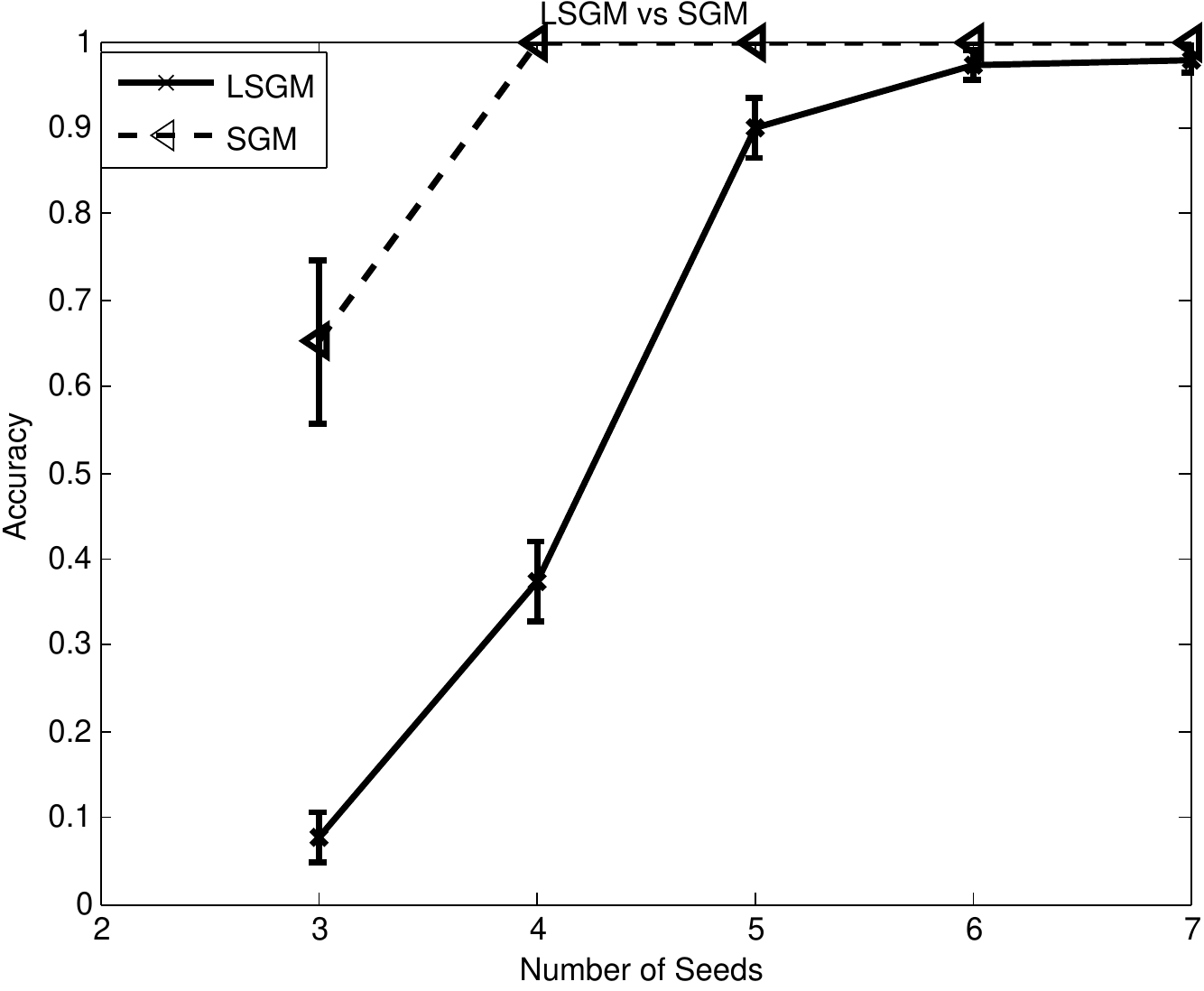}
\caption{The fraction of the unseeded vertices correctly matched across SBMs with $K=3$ blocks, block--block connectivity as specified in the text, $\vec{n}=(200,200,200)$, $\rho=.7$,  and $s=3,4,5,6,7$ seeds randomly assigned to one of the three blocks.  The dashed curve plots the fraction of unseeded vertices correctly matched by the SGM algorithm across the various $s$, with error bars $\pm 2 s.e.$  The solid curve plots the fraction of unseeded vertices correctly matched by the LSGM algorithm across the various $s$, with error bars again $\pm 2 s.e.$   Here SGM is the algorithm of \cite{FAP} run without clustering.}
\label{fig:sgmlsgm}
\end{figure}
Our two step approach first embeds and clusters the two graphs and then matches them accordingly.  Theoretically, we can embed and cluster and then match the graphs perfectly, but we next explore how much accuracy is practically lost because of the embedding step.  
When $n$ is small (e.g. $\leq 1500$) and the SGM algorithm of \cite{FAP} can be feasibly run without first clustering, the SGM algorithm will outperform LSGM in general, even in the SBM setting.  Indeed SGM utilizes the across cluster connectivity structure in the matching task, information which 
LSGM does not utilize when matching across clusters.  It is also clear that SGM is utilizing more of the information contained in the seeding than LSGM.  If the latent positions generating the SBMs are separated enough (as at assumption {\it i.} of Theorem \ref{T}) and $n$ is large enough for the clustering to be consistent across the graphs, then we will illustrate that LSGM performs excellently.  However, even in the case of perfect clustering, LSGM still needs (modestly) more seeds than SGM to achieve comparable performance.  We illustrate this in Figure \ref{fig:sgmlsgm}.  We match across two $\rho=0.7$-correlated SBMs with $K=3$ blocks, $\vec{n}=(200,200,200)$, with block--block adjacency probabilities dictated by the matrix
$$\begin{pmatrix} 0.6& 0.3 &0.2\\ 0.3&0.7&0.3\\0.2&0.3&0.7\end{pmatrix},$$
and seed values ranging from $s=3,4,5,6,7$ drawn uniformly from the 600 vertices.  The dashed curve plots the fraction correctly matched by the SGM algorithm across the various $s$, with error bars $\pm 2 s.e.$  Analogously, the solid curve plots the fraction correctly matched by the LSGM algorithm across the various $s$, with error bars again $\pm 2 s.e.$   Note that with only 4 seeds, SGM perfectly matches across the graphs, though LSGM requires 7 seeds for comparable performance.  

\begin{figure}[t!]
\hspace{35mm}
\includegraphics[width=.6\textwidth]{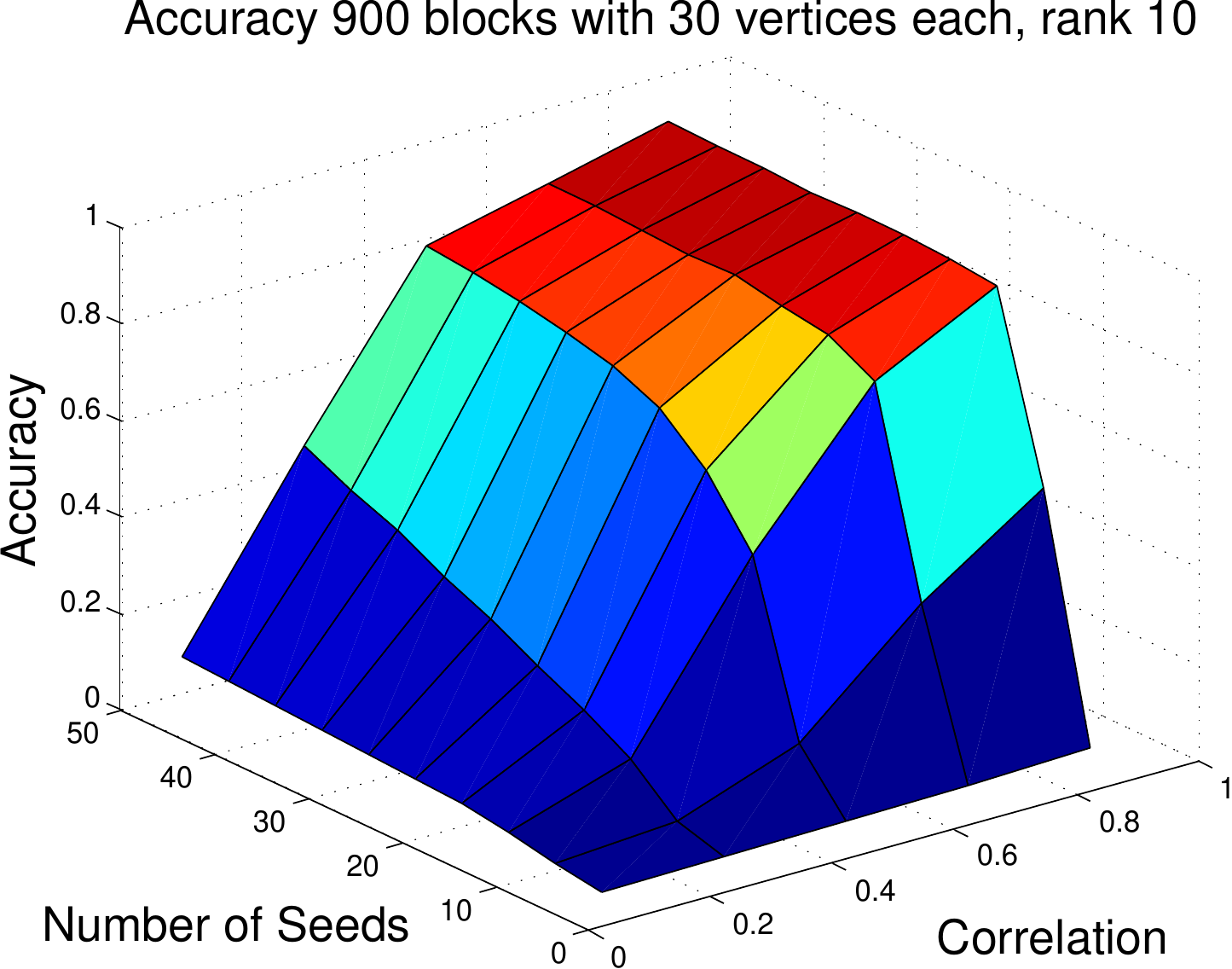}
\label{fig:sub6}

\caption{Fraction of unseeded vertices correctly matched across two $K=900$ block, $\vec{n}=30\cdot\vec{1}$, $d=10$ dimensional $\rho$-correlated SBM's with $s$ seeds drawn uniformly at random from the 27000 vertices.  Note that for each combination of $s$ and $\rho$, we ran 25 MC simulates.  All standard deviations are $<.03$ except with 10 seeds where the s.d.\@ is $0.0694, 0.2325,  0.1958$ for $\rho=0.5,0.7,0.9$.}\label{cliff}
\end{figure}
Given a consistent clustering of the graphs, LSGM needs modestly more seeds to perform as well as the full SGM.  In contrast, if the clustering is not consistent, LSGM cannot hope to match the clusters exactly.
However, while SGM can only match graphs of order $\approx 1000,$ LSGM can be used to match much larger graphs.  We demonstrate this in the following experiment, where we match two large SBM graphs.  In Figure \ref{cliff}, we plot the average accuracy of LSGM in matching the unseeded vertices in 25 MC simulations across two $K=900$ block, $\vec{n}=30\cdot\vec{1}$, $d=10$ dimensional, $\rho$-correlated SBM's with $s$ seeds drawn uniformly at random from the 27000 vertices.  The $K$ latent positions $X$ are sampled uniformly from the $d$-dimensional simplex, and we utilize the $k$-means clustering algorithm ($k$ an estimate of $K$) in Step 5 of Algorithm \ref{alg:emcl}. Note how few seeds are needed to ensure good performance for even modestly correlated graphs.  For example, we correctly match 78.75$\%$ of the unseeded vertices correctly with only 50 seeds and $\rho=0.5$.  This again reflects the consistency of our clustering procedure, and the applicability of our procedure in matching real data graphs, which are (at best) modestly correlated and have (at best) a modest number of seeds.

We do not assume knowledge of the true $K$ in the above procedure, instead estimating an appropriate $k$ from the data.  The figure shows that the matching is robust to this estimation.  We also do not assume knowledge of the true $d$, and here we used the automated spectral procedure of \cite{scree} to estimate the embedding dimension $d$.  The model is relatively low rank, and for higher rank SBM's we see slower algorithmic performance in general.  

\subsection{Scalability}
\label{scalability}
Our divide and conquer algorithm essentially is composed of four steps: embed, Procrustes, cluster, match.  The final matching step lends itself to parallelization, and insomuch as the embedding, Procrustes and clustering are computationally less expensive than the subsequent matching step, we expect our algorithm to scale well.  
Note that we observed this scaling previously in Section \ref{sec:k} as well, where we saw that on a 1600 vertex simulated graph our parallelization procedure was able to achieve an  8x  improvement in speed at minimal accuracy degradation by increasing the number of clusters and hence the number of cores that were used.
\begin{figure}[t!]
        \hspace{-10mm}
        \begin{subfigure}[b]{0.55\textwidth}
                \includegraphics[width=\textwidth]{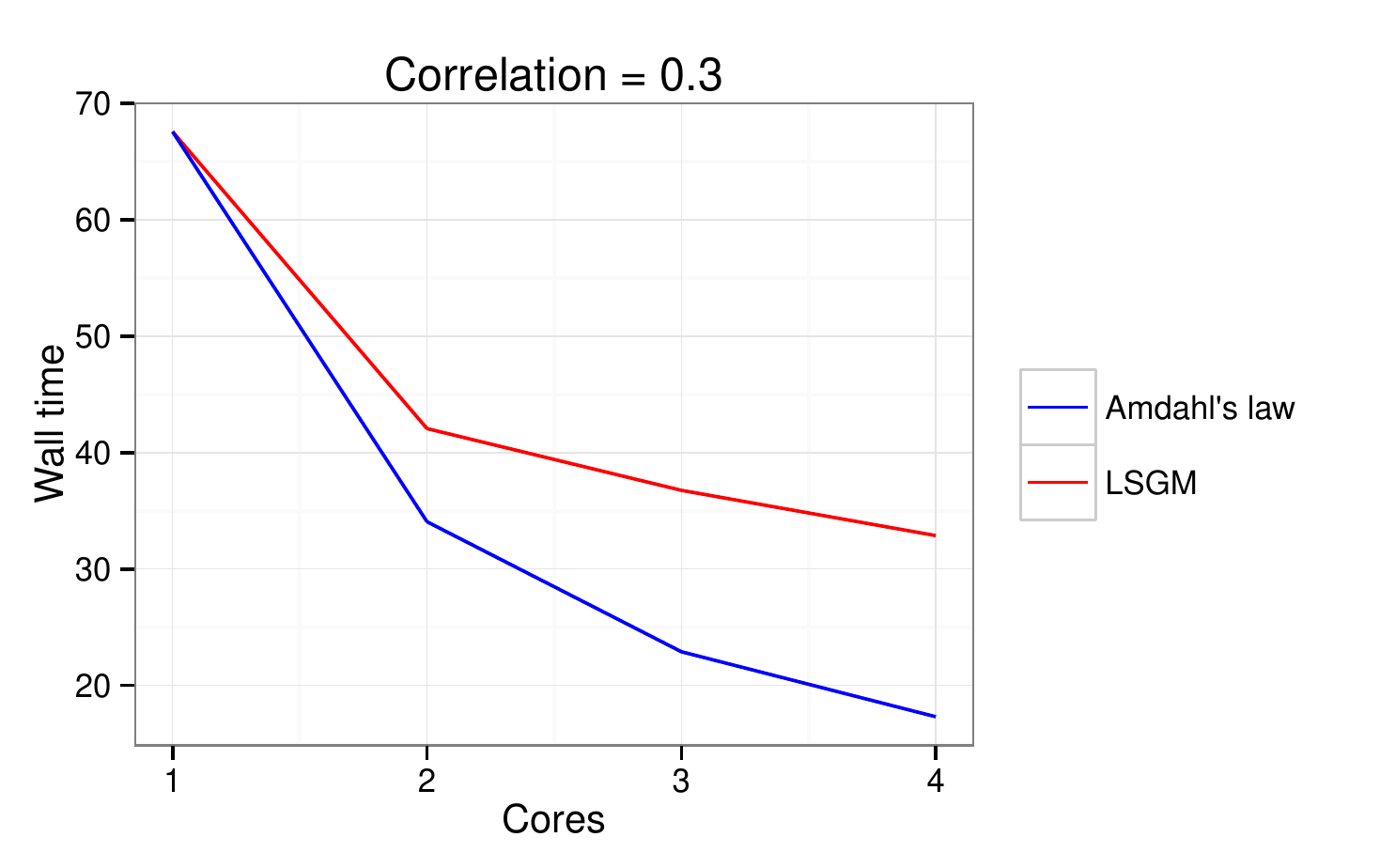}
                \caption{correlation$=$0.3}
                \label{fig:scal3}
        \end{subfigure}%
        ~ 
        \begin{subfigure}[b]{0.55\textwidth}
                \includegraphics[width=\textwidth]{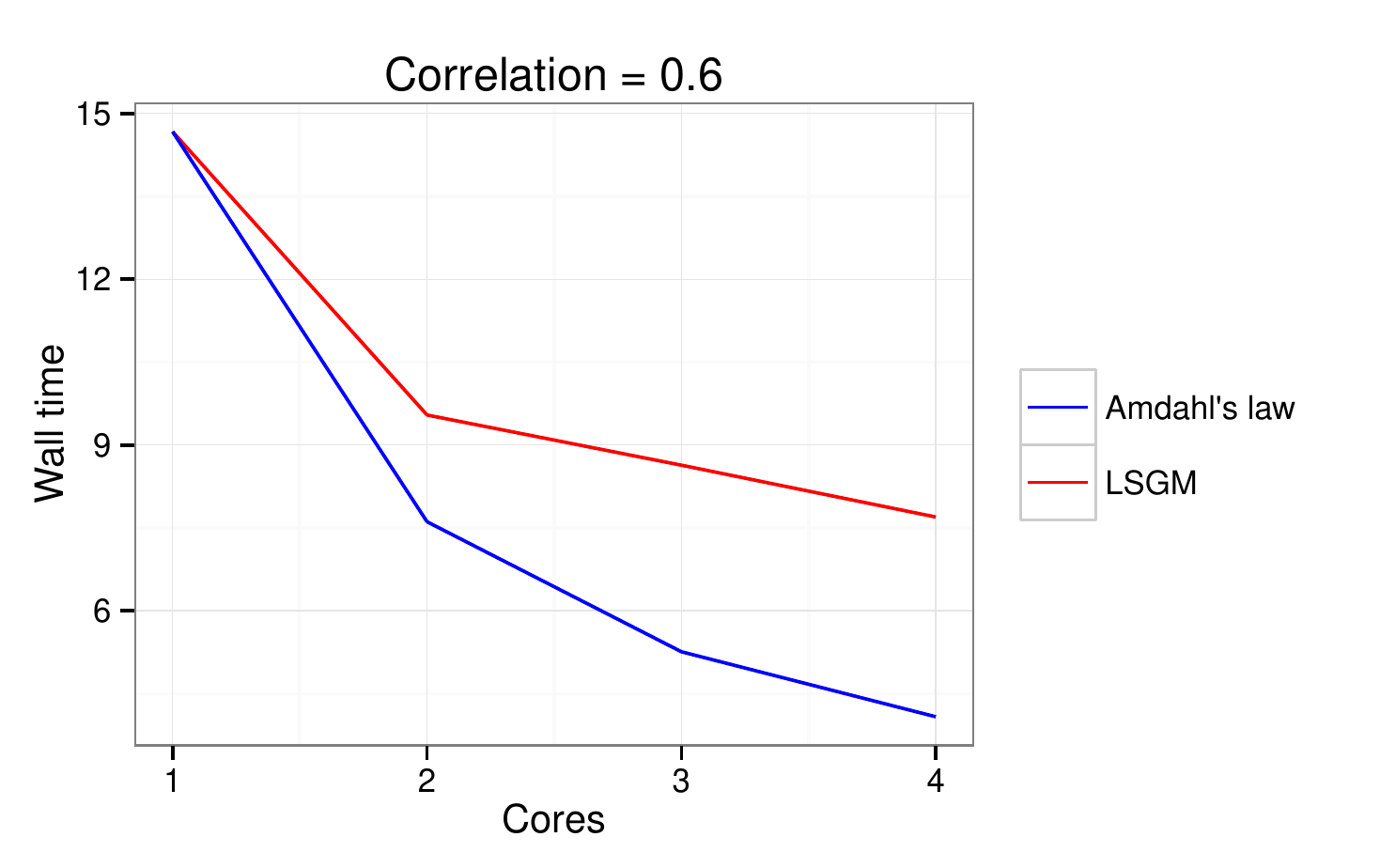}
                \caption{correlation$=$0.6}
                \label{fig:scal6}
        \end{subfigure}

        ~ 
        \hspace{-10mm}\begin{subfigure}[b]{0.55\textwidth}
                \includegraphics[width=\textwidth]{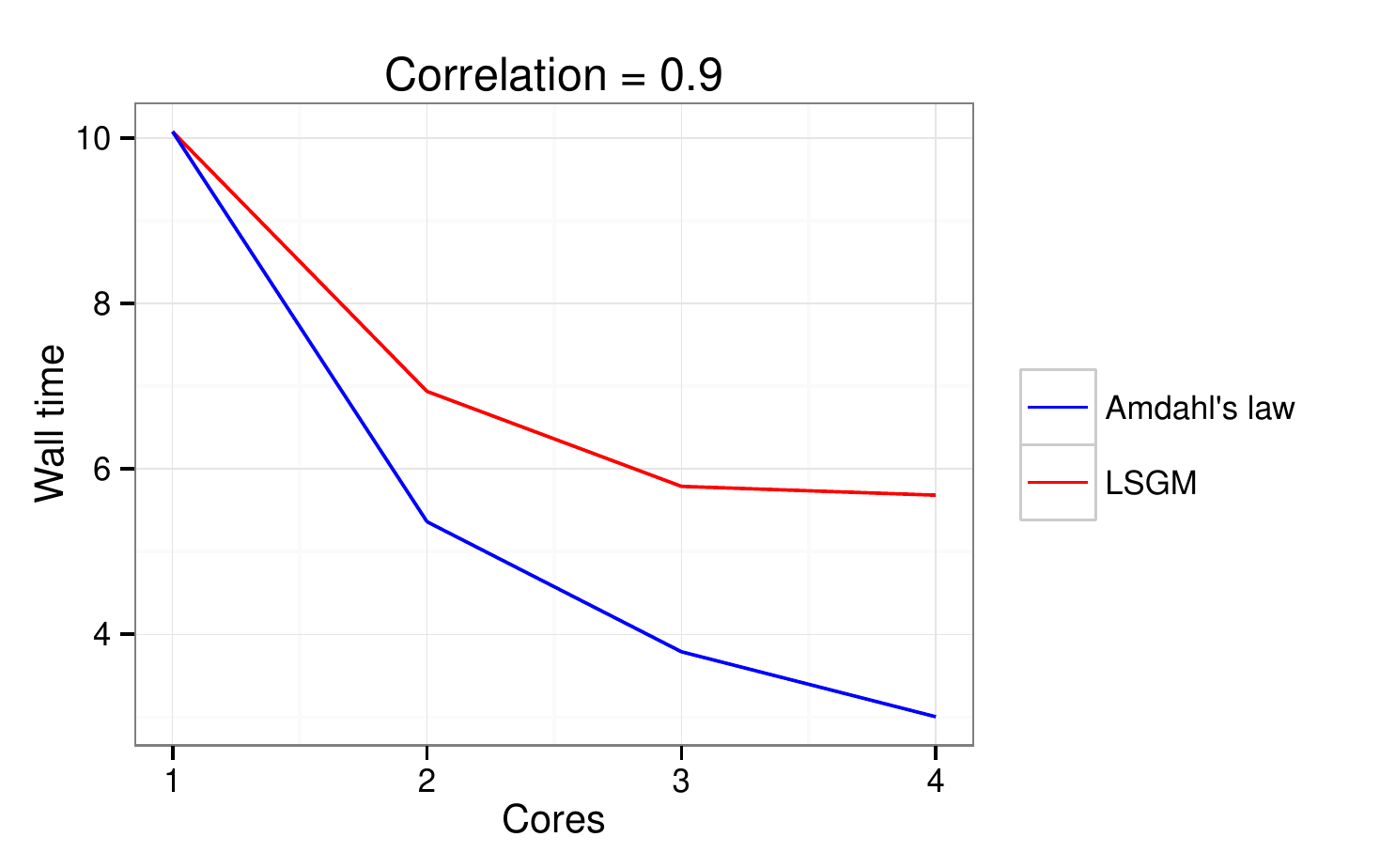}
                \caption{correlation$=0.9$}
                \label{fig:scal9}
        \end{subfigure}~\begin{subfigure}{1\textwidth}\vspace{-75mm}
\begin{tabular}{|c |c|}
\multicolumn{2}{c}{{\em Mean runtime for SGM without}}\\
\multicolumn{2}{c}{{\em first embedding or clustering}}\\
\hline
\hline
Correlation & Mean Runtime\\
\hline 
0.3    & 658.26\\
0.6    & 134.89\\
0.9    & 116.30\\ \hline
\end{tabular}
\end{subfigure}

        \caption{Runtimes when running LSGM and SGM (without first embedding and clustering the vertices) using 1,2,3,4 cores to match two SBM random graphs with 8 blocks each of size 200 (with intrablock connection probability $0.6$ and interblock connection probability $0.3$). 
        Note, SGM ran on a single core only.  
        For each experiment and each combination or $\rho$ and core number, we run 200 MC replicates, and we ran 20 MC replicates for the SGM experiment. For the full LSGM procedure, we then plot the achieved runtime against the theoretical maximum speedup possible when parallelizing as predicted by Amdahl's law.}
        \label{fig:scale}
\end{figure}

To explore this further, we run our algorithm on three pairs of SBMs with varying $\rho=0.3,0.6,0.9$.  Each SBM has 8 blocks (with intrablock connection probability $0.6$ and interblock connection probability $0.3$) each of size 200, and we run our LSGM procedure with 20 seeds utilizing 1--to--4 cores and, in all cases, clustering the graphs into 8 clusters.  We plot the resulting algorithmic wall times in Figure \ref{fig:scale} (run on a Genuine Intel laptop: model name: Intel(R) Xeon(R) CPU E31290 @ 3.60GHz with 4 processors).  
We note that with lower correlation, matching is the most costly step in our procedure as expected, while in the high correlation setting ($\rho=0.9$), the matching steps are relatively inexpensive.
In all cases, we see roughly a 2x speedup in our procedure when utilizing 4 cores. 
We lastly note that matching these graphs using SGM with $\rho=0.9$ without first embedding and clustering the graphs has average runtime $\approx116$ seconds ($\approx134$ seconds when $\rho=0.6$ and $\approx658$ seconds when $\rho=0.3$), compared with $\approx 5$ seconds ($\approx 7$ seconds when $\rho=0.6 $and $\approx 32$ seconds when $\rho=0.3)$ with our divide-and-conquer procedure using 4 cores.  See figure \ref{fig:scale} for detail.
\begin{table}[t!]
\centering

        \begin{tabular} {|c |cccc|}

\multicolumn{5}{c}{{\em Runtime in seconds}}\\ \hline

\hline
Cores & Embed & Procrustes            & Cluster              & Match   \\
\hline

\hline
& \multicolumn{4}{|c|}{ $\rho=0.3$}\\ \hline
1     & 0.53  & 0.89$\times 10^{-3}$  & 0.17$\times 10^{-2}$ & 67 \\
2     & 0.54  & 0.73$\times 10^{-3}$  & 0.18$\times 10^{-2}$ & 41 \\
3     & 0.54  & 0.72$\times 10^{-3}$  & 0.18$\times 10^{-2}$ & 36 \\
4     & 0.54  & 0.72$\times 10^{-3}$ & 0.15$\times 10^{-2}$ & 32 \\ \hline

\hline
& \multicolumn{4}{|c|}{ $\rho=0.6$}\\ \hline
1     & 0.53  & 0.89$\times 10^{-3}$  & 0.19$\times 10^{-2}$ & 14  \\
2     & 0.53  & 0.73$\times 10^{-3}$  & 0.18$\times 10^{-2}$ & 8.9 \\
3     & 0.54  & 0.72$\times 10^{-3}$  & 0.19$\times 10^{-2}$ & 8.0 \\
4     & 0.54  & 0.73$\times 10^{-3}$  & 0.19$\times 10^{-2}$ & 7.1 \\ \hline

\hline

& \multicolumn{4}{|c|}{ $\rho=0.9$}\\ \hline
1     & 0.53  & 1.06$\times 10^{-3}$  & 1.06$\times 10^{-2}$ & 9.4 \\
2     & 0.53  & 0.74$\times 10^{-3}$  & 0.20$\times 10^{-2}$ & 6.3 \\
3     & 0.54  & 0.73$\times 10^{-3}$  & 0.21$\times 10^{-2}$ & 5.2 \\
4     & 0.54  & 0.72$\times 10^{-3}$  & 0.20$\times 10^{-2}$ & 5.1 \\ \hline
\end{tabular}
\vspace{12pt}

        \caption{The table show mean runtimes when running LSGM using 1,2,3,4 cores to match two SBM random graphs with 8 blocks each of size 200 (with intrablock connection probability $0.6$ and interblock connection probability $0.3$).  For each experiment and each combination or $\rho$ and core number, we run 200 MC replicates.
        The table shows how the runtime breaks down into the four steps of the algorithm: embedding, procrusties, clustering, and matching. }
        \label{table:runtimes}
\end{table}

For each of the three correlation levels and for each of 1 to 4 cores, we also calculated the average runtime of each step of our algorithm: embedding, Procrustes, clustering and matching (see the Table \ref{table:runtimes} for details).  We see that matching is the most time intensive aspect of the procedure (especially in the low correlation setting), and that parallelizing the other components of our algorithm would yield incremental runtime improvements when compared to parallelizing the matching step.  While parallelizing the other components of our algorithm has been the subject of independent research, the gains in implementing these parallelization strategies are incremental in this setting, and therefore we do not pursue them here.

\subsection{Connectomes}
\label{brains}
We next demonstrate the effectiveness of the LSGM algorithm in a practical real data setting.
In this data set, for each of 21 subjects, we have two brain connectome graphs.  For each subject, the vertices in the connectome graphs correspond to voxels in the $64\times64\times64$ voxel diffusion tensor MRI brain mask.  Edges between vertices are present if there exists at least one neural fiber bundle connecting the voxel regions corresponding to the two vertices. The largest connected component (LCC) in these connectomes ranges from 20,000--30,000 vertices.  For more detail on the creation of these graphs and their utility in the neuroscience literature, see \cite{wrg1} and \cite{wrg2} and the references contained therein.  All the data can be found at \texttt{http://openconnecto.me/graphs} (note that we have spatially down-sampled each data point by a factor of four in each dimension).  

While our theory is proven in the setting of SBM random graphs, this example shows the applicability of our method in matching graphs with heavy-tailed degree distribution.  Indeed, when we plot on a log-log scale the degree sequence of two of the connectomes to be matched below, we see all three connectomes have a heavy-tailed degree distribution rather than the flat degree distribution we would expect from the SBM; see Figure \ref{fig6} for detail.  While our algorithm uncovers significant signal when matching across these connectomes, it will be useful to explore modifications to our approach for accommodating heavy-tailed degree graphs and power-law graphs.  We strongly suspect that there is significant signal in the degree distribution, with higher degree vertices being easier to correctly match that lower degree vertices, and we are presently working to theoretically verify and empirically explore the algorithmic impact of these heavy-tailed degrees.
\begin{figure}[t!]
        \centering
        \begin{subfigure}[b]{0.32\textwidth}
                \includegraphics[width=\textwidth]{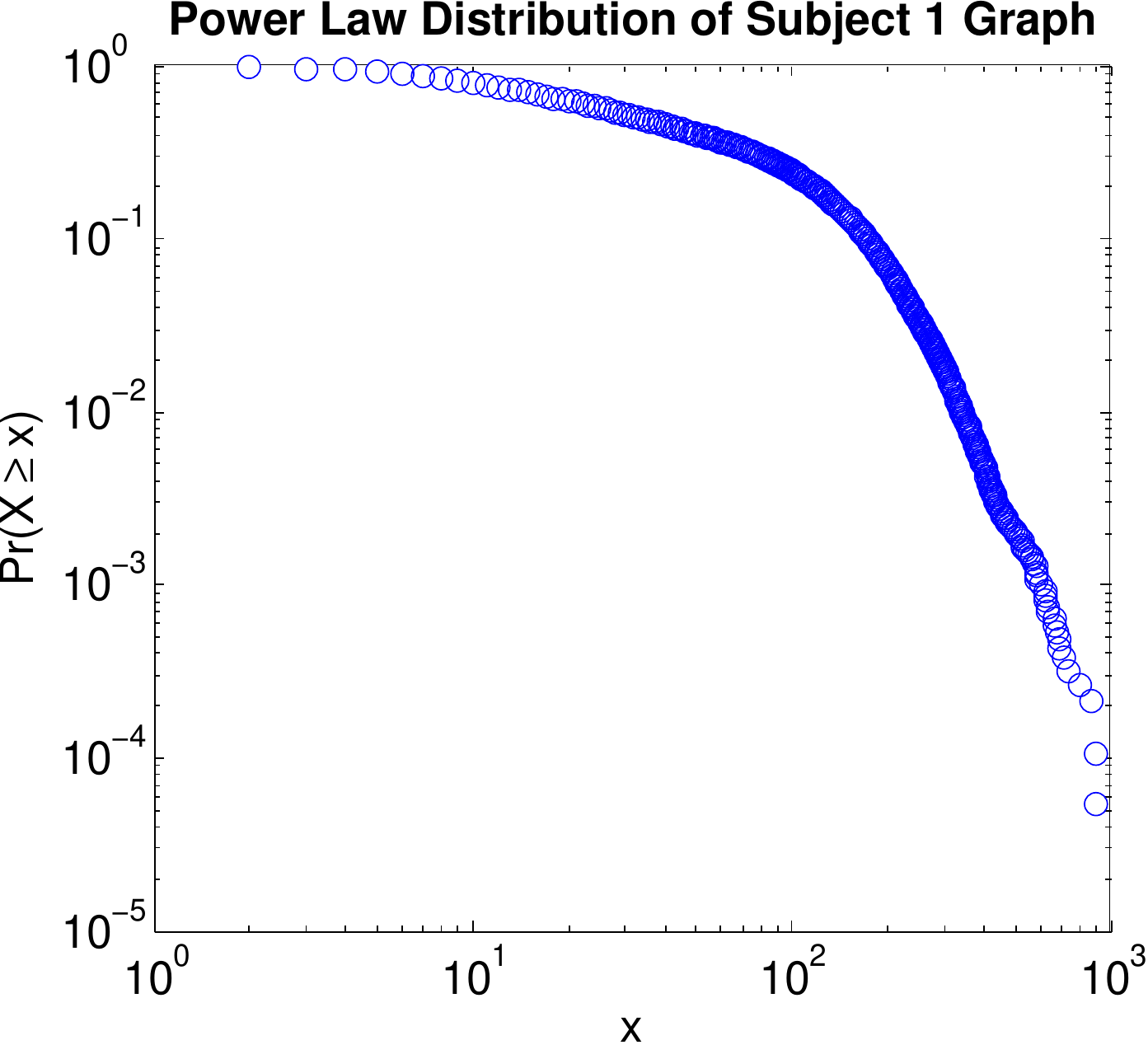}
                \caption{Graph 1}
                \label{fig:degree1}
        \end{subfigure}
        \begin{subfigure}[b]{0.32\textwidth}
                \includegraphics[width=\textwidth]{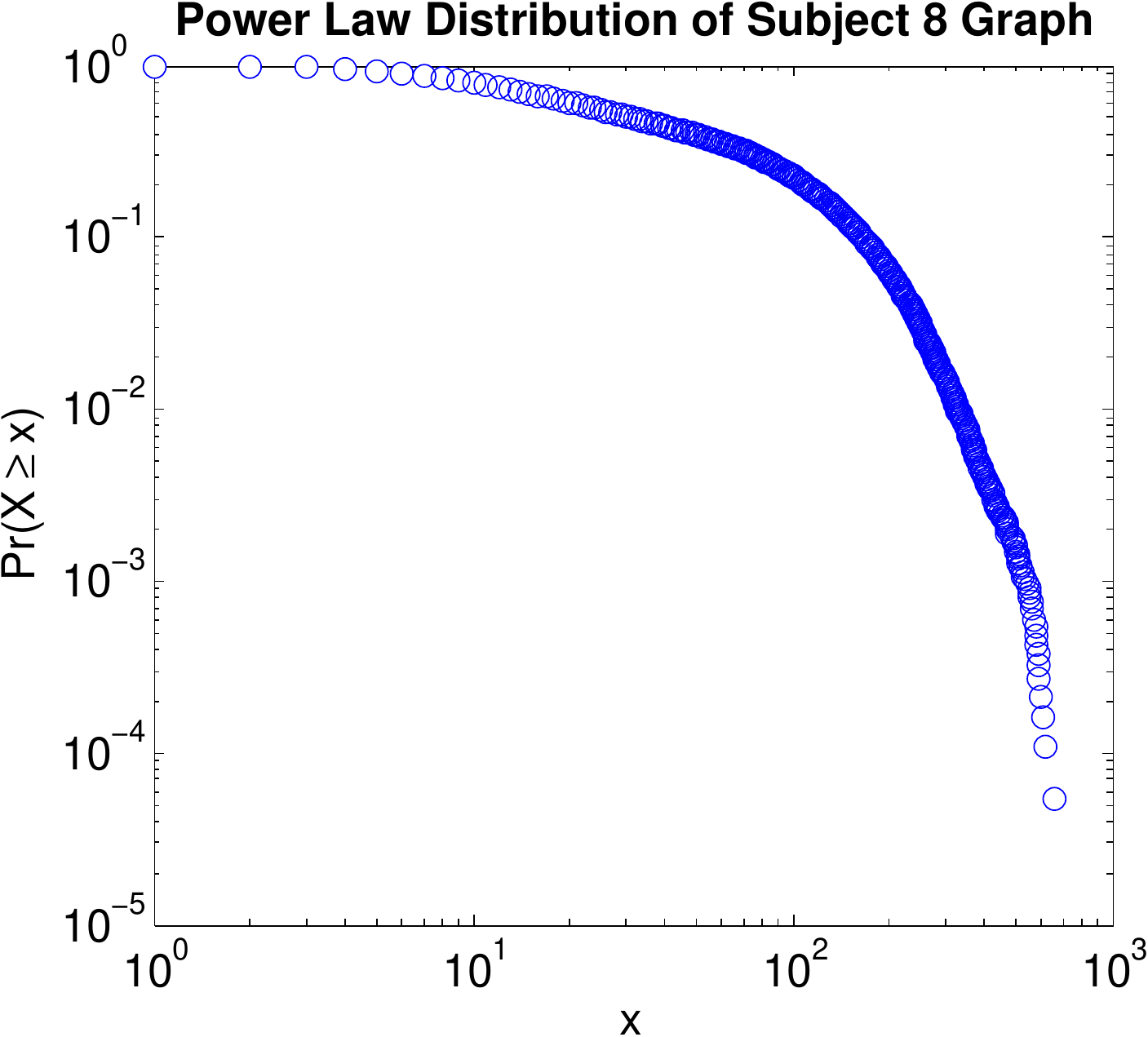}
                \caption{Graph 8}
                \label{fig:degree8}
        \end{subfigure}
        \begin{subfigure}[b]{0.32\textwidth}
                \includegraphics[width=\textwidth]{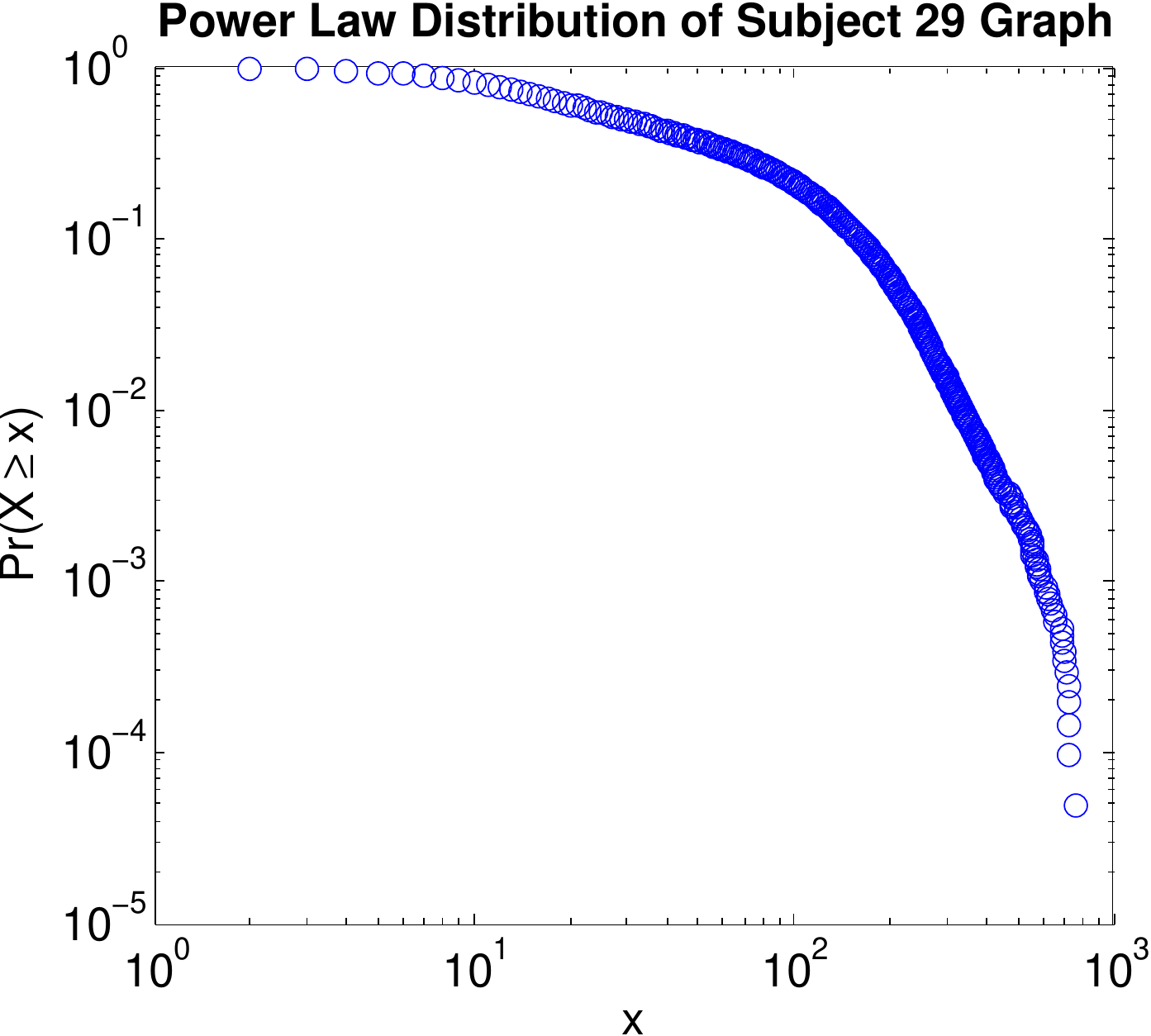}
                \caption{Graph 29}
                \label{fig:degree29}
        \end{subfigure}

        \caption{Degree distribution for connectomes 1, 8 and 29.  The degrees are plotted on a log-log scale and are strong evidence of a heavy-tailed degree distribution.}\label{fig:scal}
\end{figure}

In \cite{wrg1}, the authors collapsed the larger graphs into smaller, more manageable graphs (with vertex count $<1000$) and matched across these smaller graphs.  For any two subjects, they were able to correctly match a significantly higher percentage of the vertices for the two pairs of within--subject graphs than for the four pairs of across--subject graphs.  We obtain analogous results by running the LSGM algorithm to match across the larger, less downsampled, graphs.  
The graphs are created such that the true alignment for any two graphs matches vertices comprised of the same voxels in the $64^3$ voxel brain mask.  

In Figure 4, we highlight our results for a single pair of subjects, and note that analogous results held across the data set.  In this example, the LCC of graphs 8 and 29 are of size 21,891 and 22,307 respectively, and the LCC of graph 1 is size 22,734.  We match across the intersection of the LCC's for graphs 8 and 29 (same subject, results plotted in Figure 4) and for graphs 1 and 8 (different subjects, results plotted in Figure 4).  From the SCREE plot, we estimate the optimal embedding dimension to be $d=30$ in both cases and we cluster using $k$-means, and as noted in Section \ref{cost}, we recluster any overly large clusters---here reclustering any clusters of size $\geq 800$---and hence we initially set $k=\lceil n/800 \rceil$.    It is clear from Figure \ref{fig6} that LSGM correctly matches a significantly larger proportion of vertices for the within--subject connectomes than the across--subject connectomes.  As these connectomes are too large to feasibly run SGM (or any of the bijective matching procedures other than U---which performed very poorly here), we cannot compare the performance of LSGM to the other bijective approaches here.   

On the 2 x Intel(R) Xeon(R) CPU E5-2660 0 $@$ 2.20GHz
machine using 12 cores, we display the average runtime (wall time) when matching across connectomes for the four steps of our algorithm in Table \ref{table:brain}.
\begin{table}
\centering
\begin{tabular}{|c |c|c|c|c|c|}
\multicolumn{6}{c}{{\em Runtime in seconds for connectome experiment using $k$-means clustering}}\\
\hline

\hline
subjects & seeds & projection & Procrustes & clustering & matching \\ \hline

\hline
01-08 & 200  &  562.82 & 1.57 & 13.49 &  473.46 \\
01-08 & 1000 &  754.84 & 2.44 & 15.79 &  883.58 \\
01-08 & 2000 &  862.43 & 2.82 & 15.14 & 1495.26 \\
01-08 & 5000 &  981.86 & 3.60 & 13.55 & 2698.32 \\
08-29 & 200  &  777.11 & 1.94 & 18.16 &  569.97 \\
08-29 & 1000 &  987.08 & 2.80 & 18.47 & 1019.73 \\
08-29 & 2000 & 1096.84 & 3.26 & 19.29 & 1592.76 \\
08-29 & 5000 &  890.89 & 2.90 & 15.19 & 2902.19 \\ \hline
\end{tabular}
\caption{Runtime for LSGM on the connectome graphs.  For each of the four steps of our procedure and each combination of seeds and connectomes, we display the average wall time measured in seconds.
The clustering step is the traditional $k$-means, not the $sk$-means modification.
Again, matching is the most time intensive step.  It is interesting to note that a longer matching runtime corresponds to better algorithmic performance.  Note that the graphs are projection into $\mathbb{R}^{30}$.}
\label{table:brain}
\end{table}
Although the projection step takes longer to run than matching in some of the examples, this is an artifact of the full parallelization of the matching step; indeed, the matching step would be computationally unwieldy without parallelizing.  We also note that slower matching corresponds to better algorithmic performance.  With this in mind, we expect greater improvement from implementing our algorithm on more specialized computational hardware (and paralellizing the SVD calculation for very large graphs) and from employing hot restarts when the algorithm terminates quickly.  We emphasize that even very large graphs can be reasonably matched with a simple computing cluster.

It is worth noting that in this example (and across the entire data set), more seeds corresponded to a significantly better matched ratio for both the within--subject and across--subject pairs of graphs. 
However, for the larger values of $s$ ($s=1000,\, 2000,\, 5000$), we are unable to run the SGM subroutines utilizing the full seeding.  Instead, we used the active seed selection algorithm of Section \ref{cost} to pick an ``optimal", computationally feasible set of seeds to use in matching across each cluster.  
In all cases, our algorithm performs significantly better than chance (chance here being $1/(n-s)=[5.35e-5, 5.65e-5, 5.99e-5, 7.3e-5]$ for the 1-8 pair and $1/(n-s)=[4.87e-5, 5.12e-5, 5.39e-5, 6.43e-5]$ for the 8-29 pair for $s=[0,1000,2000,5000]$).

We also explore the potential for increased performance in LSGM by utilizing different clustering procedures.  
The brain graphs are very sparse, and there is precedent in the literature that first projecting the latent positions onto the sphere and then clustering the graphs via $k$-means results in better clustering performance in the presence of graph sparsity \cite{perfectclust}.
We call this variant of $k$-means the {\it spherical $k$-means} ($sk$-means) algorithm, and
we see that replacing standard $k$-means with $sk$-means significantly increases the performance of the LSGM algorithm.  
This result reinforces the idea that, in practice, the clustering procedure should be chosen to leverage the signal present in the data.

\begin{table}[t!]
\centering
\begin{tabular}{|c |c|c|c|c|c|}
\multicolumn{5}{c}{{\em Mean matching accuracy and standard error}}\\
\hline

\hline
& SGM&
FAQ&
Umeyama&
PATH CR \\ \hline
\hline
\text{Mean accuracy}&0.0773&0.0064&0.0034&0.0091 \\
\text{Standard Error} & 1.72e-03 & 4.25e-04 & 1.12e-04& 2.91e-04\\ \hline
\end{tabular}
\caption{The matching accuracy and standard error for matching the 8-29 within-subject pair across matching algorithms in the divide-and-conquer paradigm.  The max cluster size is set to 800, $s=2500$, the graphs are embedded into $d= 30$, and the number of Monte Carlo replicates is 20.  Note that SGM greatly outperforms the other algorithms.}
\label{table:brain2}
\end{table}
Our results reconfirm that variability in the estimated connectivity is greater between subjects than within subjects.
The estimated connectivity varies due to both noise in the collection of raw scan data and the use of a suite of pre-processing tools used to clean, register and analyze the raw data.
As a result, large scale graph matching can serve as another tool to assess the reliability of these methods. 
Furthermore, this suggests that when registering two scans from the same subject, jointly using geometric properties and connectivity will improve registration accuracy.

We lastly note that within cluster matching using SGM also significantly outperforms the other graph matching algorithms (applied post embedding and clustering) when matching across brain graphs; see Table \ref{table:brain2} for the matching accuracy and standard error (over 20 Monte Carlo replicates) for matching the 8-29 within-subject pair in the divide-and-conquer paradigm using $s=2500$ seeded vertices and embedding the graphs into $d= 30$.  
We did not run PATH and GLAG here due to scalability concerns.
We lastly note that running even the fastest of these algorithms, Umeyama's spectral matching procedure, without first performing the embedding and clustering is prohibitively slow.  Indeed, here Umeyama's algorithm has a runtime in excess of 50 hours and using over 30GB of RAM, reinforcing the necessity of the divide-and-conquer step (note that in the {\it graphm} package Umeyama was downloaded from, the large graph example has 1500 vertices).

\section{Discussion}
\begin{figure}[t!]
\hspace{10mm}\begin{subfigure}{.25\linewidth}
\includegraphics[width=3.5\textwidth]{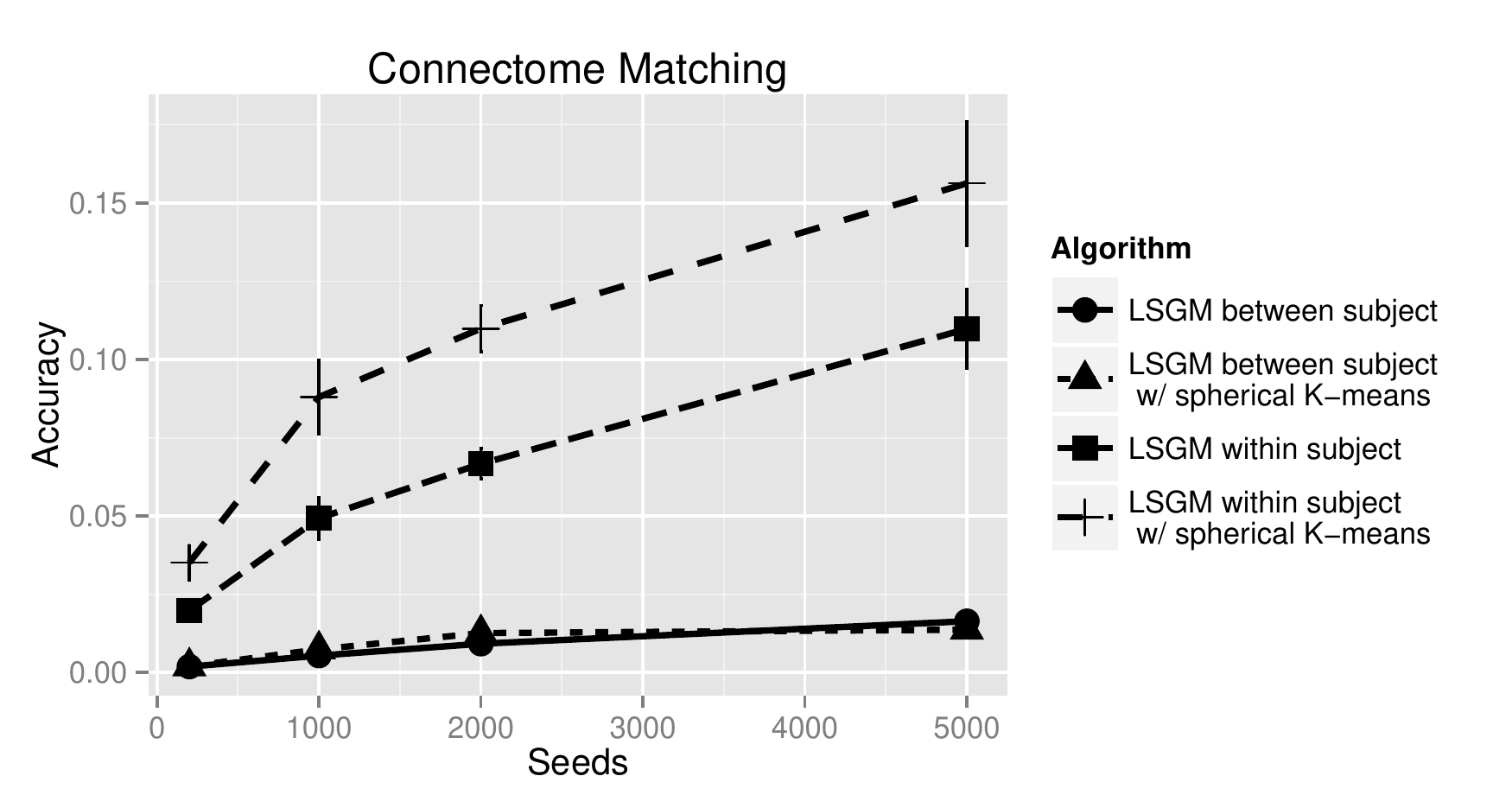}
\end{subfigure}
\caption{The fraction of the unseeded vertices correctly matched for graphs 8 and 29 (within--subject) and for graphs 1 and 8 (across--subject).  For the 8--29 pair, $n=20,541$, $d=30$.  For the 1--8 pair, $n=18,694$, $d=30$, we cluster using $k$-means and $sk$-means, reclustering any clusters of size $\geq 800$.  We plot the fraction of the vertices correctly matched in each of the four experiments for number of seeds $s=200,\, 1000,\,2000,\, $and $5000$.  Here we ran $5$ MC simulates and the error bars are $\pm 2 s.e.$}
\label{fig6}
\end{figure}
Many graph inference tasks rely on being able to efficiently match across graphs.  State--of--the--art bijective approximate graph matching algorithms have computational complexity $O(n^3)$---rendering them infeasible (without significant computational resources) for very large graphs.  We present the fully parallelizable LSGM approximate graph matching algorithm which, under some mild conditions, has computational complexity $O(n^2d)$---a marked improvement over $O(n^3)$.
We demonstrate, via simulated data examples and a real data example, the effectiveness of our LSGM algorithm in performing seeded graph matching across large graphs, which heretofore were unassailable using existing bijective matching techniques.  In addition, we theoretically justify our divide-and-conquer procedure in the SBM regime by proving that the procedure perfectly matches correlated SBM random graphs under some mild assumptions.

Our algorithm allows for flexibility in the choice of clustering and matching procedure.  We focused on $k$-means clustering here due to its ease of implementation and theoretical tractability, but the clustering procedure can (and should!) be chosen to leverage the signal present in the data.  The variety of matching procedures implemented point to the need for seeding 
the rest of the bijective graph matching procedures.

When using the seeds to match, we need to intelligently choose as many seeds as is feasible in the subsequent matching task.  We present a procedure for dynamically selecting seeded vertices.  Our procedure also provides a heuristic for defining ``good'' seeded vertices, and we are working on extending this heuristic towards the task of active learning of seeded vertices.

\bigskip

\noindent{\bf Acknowledgments:}
This work is partially supported by a
National Security Science and Engineering Faculty
Fellowship (NSSEFF), Johns Hopkins University Human Language
Technology Center of Excellence (JHU HLTCOE), and the
XDATA program of the Defense Advanced Research Projects
Agency (DARPA) administered through Air Force Research Laboratory
contract FA8750-12-2-0303. We also would like to thank William Roncal Gray, R. Jacob Vogelstein and Disa Mhembere for their help with the connectome data and thoughtful discussions and suggestions.

 \end{document}